\documentclass[runningheads]{llncs}

\usepackage{eccv}

\usepackage{eccvabbrv}

\usepackage{graphicx}
\usepackage{booktabs}
\usepackage{enumitem}
\usepackage{pifont}
\newcommand{\cmark}{\ding{51}}%
\newcommand{\xmark}{\ding{55}}%
\usepackage[accsupp]{axessibility}  
\newcommand{\eric}[1]{\textcolor{black}{#1}}
\newcommand{\ourmodel}{LucidDreaming\xspace}

\usepackage{hyperref}

\usepackage{orcidlink}
\usepackage{wrapfig}

\begin{document}

\title{LucidDreaming: Controllable Object-Centric 3D Generation} 


\author{
    Zhaoning Wang\inst{1} \and
    Ming Li\inst{1} \and
    Chen Chen\inst{1}\orcidlink{0000-0003-3957-7061}
}

\authorrunning{Z.~Wang et al.}

\institute{Center for Research in Computer Vision, University of Central Florida}

\maketitle

\begin{abstract}
\eric{With the recent development of generative models, Text-to-3D generations have also seen significant growth, opening a door for creating video-game 3D assets from a more general public. Nonetheless, people without any professional 3D editing experience would find it hard to achieve precise control over the 3D generation, especially if there are multiple objects in the prompt, as using text to control often leads to missing objects and imprecise locations. 
In this paper, we present LucidDreaming as an effective pipeline capable of spatial and numerical control over 3D generation from only textual prompt commands or 3D bounding boxes. Specifically, our research demonstrates that Large Language Models (LLMs) possess 3D spatial awareness and can effectively translate textual 3D information into precise 3D bounding boxes. We leverage LLMs to get individual object information and their 3D bounding boxes as the initial step of our process. Then with the bounding boxes, We further propose clipped ray sampling and object-centric density blob bias to generate 3D objects aligning with the bounding boxes. We show that our method exhibits remarkable adaptability across a spectrum of mainstream Score Distillation Sampling-based 3D generation frameworks and our pipeline can even used to insert objects into an existing NeRF scene. Moreover, we also provide a dataset of prompts with 3D bounding boxes, benchmarking 3D spatial controllability. With extensive qualitative and quantitative experiments, we demonstrate that LucidDreaming achieves superior results in object placement precision and generation fidelity compared to current approaches, while maintaining flexibility and ease of use for non-expert users. \textbf{Videos are available at: \url{https://www.zhaoningwang.com/LucidDreaming/}}}

\end{abstract}

\begin{figure*}[t]
\begin{center}
\includegraphics[width=\linewidth]{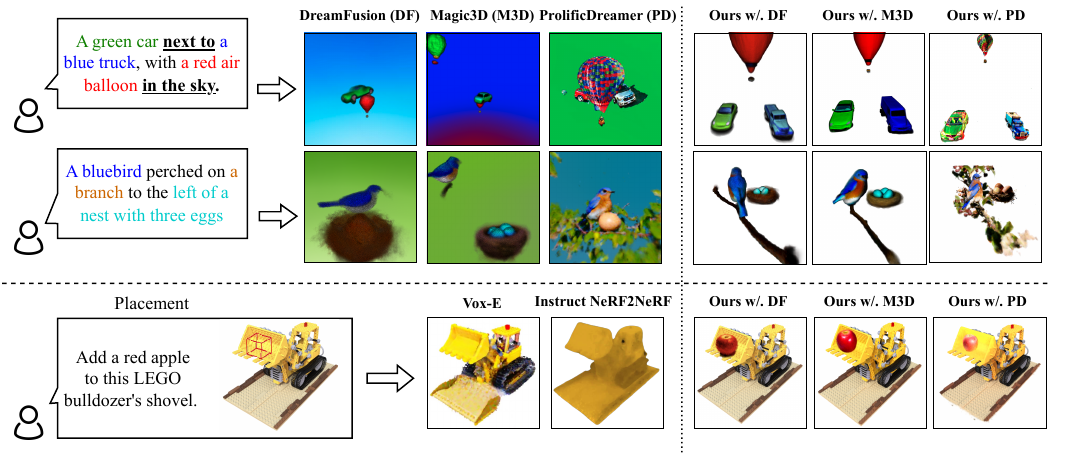}
\captionof{figure}{ 
\eric{Our pipeline enables numerical and spatial controls with just textual information, while using baseline methods\cite{poole2022dreamfusion, lin2023magic3d,wang2023prolificdreamer} only would often fail to fully capture the controlling logic (top). At the bottom, we show an application of our pipeline generating and placing specified objects with a NeRF with user-provided bounding boxes and prompts. This is typically difficult due to NeRF's implicit representation.}
} 
    \label{fig:teaser}
    \end{center}%
\end{figure*}

\section{Introduction}
\label{sec:intro}

With the proliferation of gaming and media industries, 3D content creation has become a crucial element in their development pipelines, ensuring immersive and visually captivating experiences for users. Amidst the notable progress in text-to-image diffusion models~\cite{rombach2022high, ho2020denoising}, 3D content generation has seen significant enhancements~\cite{poole2022dreamfusion, wang2023prolificdreamer, chen2023it3d} with the introduction of Score Distillation Sampling (SDS)~\cite{poole2022dreamfusion}, which distills knowledge from 2D diffusion models into a Nerual Radiance Field (NeRF)~\cite{mildenhall2021nerf}\eric{, a popular 3D representation}. Such progression empowers individuals to instantiate intricate 3D models from mere textual descriptions. 

Though it is a great way to create 3D content in spectacular detail, controlling the generation is not a trivial task. \eric{Generating and positioning multiple objects often require iterative processes and specialized 3D software skills, such as model editing, limiting complex asset creation to professionals rather than the general public.} One straightforward way of exerting control would be using the text prompt directly in the 3D generation, such as \textit{``6 apples arranged in a 3 by 2 grid"}. Nonetheless, diffusion models are deficient in comprehending such underlying logic in the prompts, especially with spatial relations or numbers~\cite{lian2023llm, huang2023t2i}. Furthermore, the distillation of 2D models handles occlusions poorly. For example, the SDS will always try to adhere to ``six apples" in every single view in the former prompt, contributing to missing/extra objects and wrong positions. As shown by the generative baselines in Fig.~\ref{fig:teaser} (top-left), objects are either omitted or fused into composite entities when using naive control from the prompts. Consequently, relying solely on textual input to steer the 3D generative process proves to be quite impractical under the current context.

To overcome these constraints, recent methods have implemented controllability from a compositional standpoint, utilizing conditional diffusion models to constraint generation through the projection of 3D boxes within 2D images~\cite{po2023compositional, shum2023language}. 
However, such approaches can not adapt to enhanced diffusion models in a plug-and-play manner without finetuning.
We also observe that the 3D bounding boxes used in the current work~\cite{po2023compositional} have to be connected, i.e., all boxes have to share at least one common face with the others. This would prevent such methods from generating objects with distinctness.
\eric{In this paper, we aim to create a pipeline that unlocks the possibility for people with no professional experience to create more complex multi-objects while also adaptive to different popular SDS-based 3D asset creation methods.}

To this end,  we propose \textbf{\ourmodel} as a straightforward yet effective method to solve the problem. 
To begin with, previous work fits the NeRF scenes globally, rendering each view as separate images and only enforcing conditions on the 2D images individually. Manipulating only in 2D image space would lead to inaccuracies in text-to-3D frameworks or the need for conditional diffusion models to maintain control. To resolve such limitations, we propose clipped ray sampling, enabling isolated object rendering in 3D space, and thereby enhancing controllability significantly.

However, the naive integration of clipped ray sampling does not solve the omission or clustering of objects in the baseline methods. Such issues arise due to the common use of uni-sphere density blob initialization in 3D generation~\cite{poole2022dreamfusion,lin2023magic3d}, predisposing towards single-entity creation. In this light, we propose object-centric density initialization, further promoting distinctiveness between generated objects. Moreover, to achieve an end-to-end methodology, we integrate a Large Language Model, enabling control over generation through a simple prompt that translates into bounding boxes.

\begin{figure*}[t]
    \centering
    \vspace{-0.1cm}
    \includegraphics[width=\textwidth]{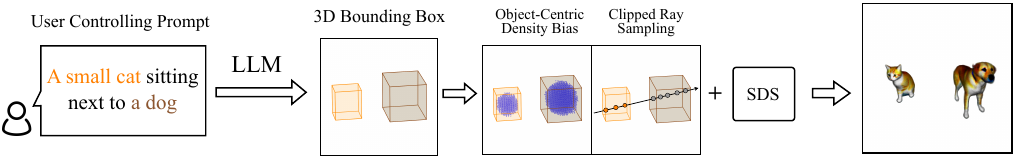}
    \caption{A high-level overview of our pipeline, controlling prompts are decomposed into 3D bounding boxes with LLMs, such as GPT4. Then in \ourmodel, object-centric density bias and clipped ray sampling are used with Score Distillation Sampling (SDS) loss to align the generation with the user’s control.}
    \vspace{-0.3cm}
    \label{fig:pipeline}
\end{figure*}

Apart from existing generation methods which only generate from scratch, our approach is also applicable to 3D content generation within a pre-trained NeRF scene, as shown in Fig.~\ref{fig:teaser} (bottom).
Alike changes to an existing scene would normally be achievable through previous 3D editing methods~\cite{instructnerf2023, sella2023vox, zhuang2023dreameditor}, but only limited to slight semantic modifications to existing objects, like adding accessories or altering object styles. Conversely, our method can generate entirely new objects within existing spaces, independent of the scene's current objects.

Fig.~\ref{fig:pipeline} illustrates a high-level overview of our framework. Due to the modular design of the method, our pipeline is compatible with a majority of SDS-based 3D generation methods, exhibiting its wide adaptability, exemplified in Fig.~\ref{fig:teaser} (right). We further provide a dataset comprising prompts alongside corresponding 3D bounding boxes and object descriptions, aimed at facilitating benchmarking in 3D generation controllability. In summary, our contributions are listed as follows:

\begin{itemize}

    \item \eric{We propose \ourmodel, a plug-and-play pipeline for generating multiple objects with spatial and numerical relationships simply from text prompts, allowing asset creation without any profession 3D software knowledge.}
    \item We present clipped ray sampling and object-centric density bias initialization, for individual object rendering and optimization. This enables controllable 3D content generation both from scratch and in pre-trained scenes. 

    \item We show our method precisely adheres to the controlling condition and is adaptive to multiple SDS-based methods. \eric{We illustrate that Large Language Models have 3D awareness and Large Multi-Models such as GPT4-V can be used to evaluate controllability. } 
    \item We also provide a dataset of scene prompts with individual 3D bounding boxes with object descriptions for benchmarking controllability.
\end{itemize}

\section{Related Work}
\label{sec:related_work}

\paragraph{2D Diffusion Models.}
With the introduction of the diffusion models~\cite{ho2020denoising, song2020denoising}, there has been significant progress in text-guided image generation, showing remarkable capabilities in creating high-fidelity image synthesis~\cite{rombach2022high, saharia2022photorealistic, ramesh2022hierarchical}. Due to the enriched semantics and elevated controllability of pre-trained text-to-image diffusion models, numerous research efforts~\cite{liu2022compositional, xie2023boxdiff, lian2023llm, zhang2023controllable} have proposed controlling image generation with conditioning priors, such as bounding boxes, to meticulously dictate the locations where objects should be generated. Nonetheless, extending 2D controlling methods to a 3D context is not a trivial task due to the increased complexity and dimension in 3D representation.

\paragraph{3D Generation with Diffusion.}
Recently, implicit 3D representations, notably Neural Radiance Fields (NeRF)~\cite{mildenhall2021nerf,barron2021mip,barron2022mip,muller2022instant} have garnered substantial popularity. The fast-growing of 2D diffusion models has facilitated multiple 3D synthesis methods with such representation. In particular, the Score Distillation Sampling (SDS) is introduced in
DreamFusion~\cite{poole2022dreamfusion} and concurrently in Score Jacobian Chaining~\cite{wang2023score} to generate 3D contents from the text. The SDS applies 2D Text-to-Image diffusion models to optimize the NeRF representations. A series of subsequent studies~\cite{tsalicoglou2023textmesh,lin2023magic3d,Chen_2023_ICCV,wang2023prolificdreamer,huang2023dreamtime,armandpour2023re,chen2023it3d,zhu2023hifa, qian2023magic123,liu2023one,shi2023mvdream,ye2023consistent,liu2023syncdreamer} based on SDS has improved the quality of outcomes significantly with revised 3D representations and sampling. However, imposing vocabulary constraints, like spatial relationship or numeracy, often results in generation failures, as diffusion models struggle to grasp the inherent reasoning~\cite{gokhale2022benchmarking,huang2023t2i}.

\paragraph{3D Controllability in Generation.}
Prior to the advent of proficient generative models, the majority of 3D manupulation research has primarily focused on object deformation~\cite{yuan2022nerf, yang2022neumesh, kania2022conerf}, animation~\cite{yang2021learning}, removal~\cite{mirzaei2022laterf}, and translation~\cite{guo2020object, yang2021learning}. Recently, with the emergence of diffusion models and SDS-based techniques, controlling 3D representations with generative capabilities has surged, enabling semantic control~\cite{ shao2023control4d}, and compositional manipulation~\cite{cohen2023set, po2023compositional, shum2023language}. However, existing methods are either limited to demanding custom diffusion intermediate~\cite{po2023compositional, shum2023language}, or requiring explicit 3D geometry priors~\cite{li2023focaldreamer, Chen_2023_ICCV, cohen2023set}. 
In contrast, our method enables precise 3D generation control using simple constraints, such as bounding boxes, instead of elaborated geometry priors. We also eliminate the need for specialized mediums like DreamBooth~\cite{ruiz2023dreambooth, shum2023language} or conditional diffusion models~\cite{po2023compositional}, allowing flexible adaptation to established SDS-based 3D generation frameworks~\cite{wang2023prolificdreamer,poole2022dreamfusion} with minimal effort.

\paragraph{3D Controlability in Editing.}
With the prevalence of NeRF representations, a wide range of research~\cite{wang2023inpaintnerf360, mirzaei2023reference, zhuang2023dreameditor,instructnerf2023, sella2023vox, fang2023text, gordon2023blended} are conducted on controlled edit with a pre-trained NeRF Scene. However, they primarily concentrate on editing a pre-existing object in the editing location, mostly focusing on adding accessories~\cite{zhuang2023dreameditor, sella2023vox, li2023focaldreamer,gordon2023blended}, removing objects~\cite{wang2023inpaintnerf360}, or altering object styles~\cite{instructnerf2023, fang2023text}. Our method complements these by focusing on the creation of new objects the void space of the NeRF scene, thereby extending the possibilities of 3D scene manipulation.

\section{Preliminary}
\label{sec:background}

\paragraph{NeRF and Instant-NGP.}

Neural Radiance Fields (NeRF)~\cite{mildenhall2021nerf} is a concise 3D scene representation through a function \(f_\theta\), utilizing a parameterized MLP (multi-layer perceptron) with adjustable parameters (\(\theta\)). Essentially, NeRF transforms a 3D coordinate (\(\mathbf{x}\)) and viewing angle (\(d\)) to density (\(\sigma\)) and RGB color (\(\mathbf{c}\)) via \(f_\theta:(\mathbf{x}, d) \rightarrow(\sigma, \mathbf{c})\).
Given a posed camera, each of its rendered image $\mathbf{I}$'s pixels determines a ray represented by an origin $\mathbf{o}$ and direction $\mathbf{d}$. Then the 3D points can be represented with the ray function:  $\mathbf{r}(t)=\mathbf{o}+t \mathbf{d}$, where $t$ is the depth of the ray. For each ray $r$ we sample $m$ points from $\{t_1, t_2,...,t_m\} \in M$ and we query the MLP and accumulate the outcome to ray color $\hat{\mathbf{C}}(\mathbf{r})$:
\begin{equation}
    \hat{\mathbf{C}}(\mathbf{r})=\sum_{i=1}^M T_i\left(1-\exp \left(-\sigma_i \delta_i\right)\right) \mathbf{c}_i,
\label{eq:NeRF_vanilla}
\end{equation}
where the distance of adjacent samples $\delta_i=t_{i+1}-t_i$ and the accumulated transmittance $T_i =\exp \left(-\sum_{j=1}^i \sigma_{j} \delta_{j}\right)$.

Instant-NGP~\cite{muller2022instant} is an optimized NeRF representation with hash encoding, enabling better performance. It is widely used by numerous 3D generation frameworks. Notably, along with the encoding, it also introduces a binary occupancy grid $G(x)$ to speed up the ray sampling with $c$ and $\sigma$ in Eq.~\ref{eq:NeRF_vanilla}:
\begin{equation}
  (\mathbf{c}, \sigma) = 
  \begin{cases} 
      f_{\theta}(\mathbf{x},d) & \text{if } G(\mathbf{x}) = 1, \\
      (0, 0)                    & \text{otherwise}.
  \end{cases}
  \label{eq:grid}
\end{equation}

In this way, the empty positions are skipped for faster sampling. The $G$ is updated every 16 steps such that $G(\mathbf{x})$ is set to 1 for all $\mathbf{x}$ where $\sigma_{\mathbf{x}} > \textit{threshold}$ ~\cite{muller2022instant}.

\paragraph{Score Distillation Sampling.}
DreamFusion~\cite{poole2022dreamfusion} proposes the SDS loss that distills the knowledge in 2D diffusion models for 3D generation. With the rendered image $\mathbf{I}$ from NeRF $\theta$, it adds Gaussian noise $\epsilon$ to $\mathbf{I}$ and applies the loss with the predicted noised $\epsilon_\phi$ from diffusion model $\phi$:
\begin{equation}
   \nabla_\theta \mathcal{L}_{\text{SDS}}(\phi, \mathbf{I})=\mathbb{E}_{\epsilon, t}\left[w(t)\left(\epsilon_\phi\left(\mathbf{I}_t ; y, t\right)-\epsilon\right) \frac{\partial \mathbf{I}}{\partial \theta}\right],
  \label{eq:sds}
\end{equation}
where $y$ would be the input text and $t$ is the level of noise.

Particularly, to guarantee centered position and efficiency, DreamFusion~\cite{poole2022dreamfusion} applies a spatial density bias  at NeRF initialization with a Gaussian probability density function:
\begin{equation}
    \sigma_{\text {init }}(\mathbf{x})=\lambda_\sigma \cdot \exp \left(-\frac{\|\mathbf{x}\|^2}{2 s_{\sigma}^2}\right),
    \label{eq:blobinit}
\end{equation}
where $s_{\sigma}$ represents the standard deviation regarding density $\sigma$ and $\mathbf{x}$ is the 3D position.
Although such initialization keeps generated 3D objects centered, it adversely affects controllability. As shown in Fig.~\ref{fig:blob} (a), this initialization creates a sphere-shape density, which leads to poor performance if objects are not exactly at the center. Moreover, combined with the occupancy grid, this would further induce gradient vanishing and no object would be generated.

\section{Method}
\label{sec:method}

We propose \ourmodel, an effective way to control 3D generation with bounding boxes. 
Specifically, we introduce clipped ray sampling in Sec.~\ref{sec:clipped-ray-sampling} to ensure the individual rendering within the controlling boxes, and object-centric density bias initialization in Sec.~\ref{sec:object-centric init} to place the objects strictly centered within their boxes.
Eventually, to make our pipeline end-to-end, we utilize a Large Language Model to decompose a complex prompt into corresponding bounding boxes and descriptions, as detailed in Sec.~\ref{sec:LLM}.

\subsection{Clipped Ray Sampling}
\label{sec:clipped-ray-sampling}

The volume rendering in NeRF processes the scene in its entirety, complicating discrete object control. To this end, we propose clipped ray sampling to achieve controllable and individual 3D object generation. Formally, a ray is represented as $\mathbf{r}(t)=\mathbf{o}+t \mathbf{d}$, with origin $\mathbf{o}$, direction $\mathbf{d}$, and depth $t$. For an object with a bounding box, we can calculate the intersections of the ray with its bounding box, $t_{entry}$ and $t_{exit}$, as the intersect points the ray gets in and out of the box with methods covered in~\cite{marrs2021ray}. Then during the volume rendering, we only keep the points with depth $t$ greater than $t_{entry}$ and less the $t_{exit}$.  Along one ray, for all $\mathbf{x} = \mathbf{o}+t_{\mathbf{x}} \mathbf{d}$, we denote color $\mathbf{c}$ and density $\sigma$ as:
\begin{equation}
      \hspace{-2pt}(\mathbf{c}, \sigma) = 
  \begin{cases} 
      f_{\theta}(\mathbf{x},d) 
        &\text{if} \: ( t_{\mathbf{x}} >t_{entry} \: \text{ and }  \:   t_{\mathbf{x}} < t_{exit} ),
 \\
      (0, 0)                    & \text{otherwise}.
  \end{cases}
\label{eq:clipped-ray-sampling}
\end{equation}

\begin{wrapfigure}{r}{0.5\textwidth}
    \centering
    \includegraphics[width=\linewidth]{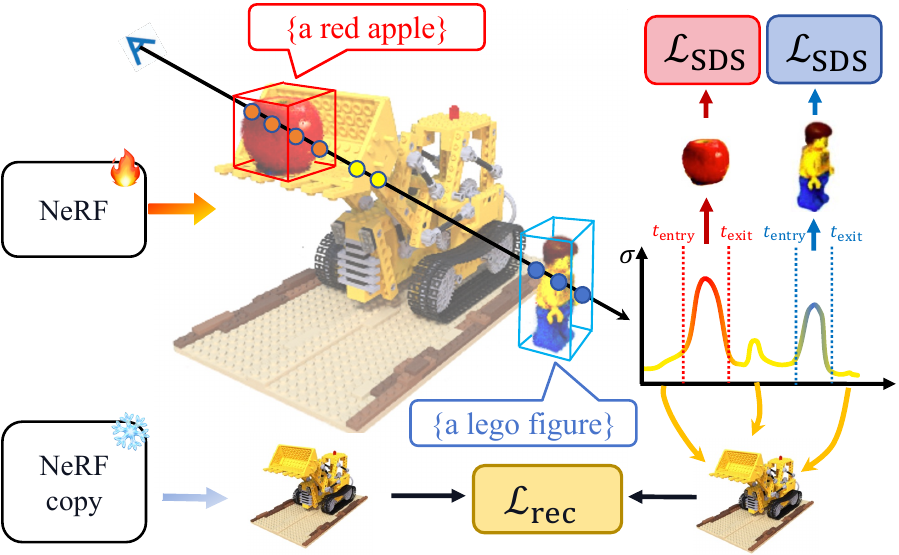}
        \caption{Clipped Ray Sampling. Points within object boxes are sampled between $t_{entry}$ and $t_{exit}$ for $\mathcal{L}{\text{SDS}}$. Outside points use $\mathcal{L}{\text{rec}}$ against frozen NeRF.}
    \label{fig:CRS}
\vspace{-0.2cm}
\end{wrapfigure}


In this way, we can perform rendering separately for each object with Eq.~\ref{eq:NeRF_vanilla} to get the final color $\hat{\mathbf{C}}(\mathbf{r})$. This enables us to easily aggregate the colors into rendered image $\mathbf{I}$ and apply SDS loss $\mathcal{L}_{\text{SDS}}^i$ separately with Eq.~\ref{eq:sds}, where $i$ is the index of objects.
By this means, we guarantee each object is rendered within its corresponding bounding box and will not go beyond the range. As exemplified in Fig.~\ref{fig:CRS}, the 3D objects of \textit{``a red apple"} and \textit{``a Lego figure"} can be optimized and generated separately with individual SDS losses.
 
\vspace{-0.3cm}
\paragraph{Scene Preservation.} While the clipped ray sampling ensures that the generation is confined to the specified bounding boxes, it simultaneously discards anything existing outside of these bounds during the sampling. This results in an unpredictable outcome from NeRF in these positions, frequently leading to floaters and noises like those shown in Fig.~\ref{fig:ablation}. Additionally, our focus extends to generative processes not solely from a blank canvas but also within the context of a pre-trained NeRF scene. To address both these aspects effectively, we employ inverse clipped ray sampling in the pipeline as a comprehensive solution.

To begin with, we keep a copy of the initialized NeRF model, either from an empty canvas or from a pre-trained scene, and freeze it throughout the generation. In the case of 3D generation, the frozen model would be a network with totally empty occupancy grid, rendering nothing but void. For the pre-trained scenes, it would be the same as the existing scene. Then during the optimization, we render all the points outside the bounding boxes with an inverse clipped ray sampling. Formally, given $n$ bounding boxes, we define inverse clipped ray sampling as:
\begin{equation}
      \hspace{-2pt}(\mathbf{c}, \sigma) = 
  \begin{cases} 
      f_{\theta}(\mathbf{x},d) 
        &\text{if} \: \left[\bigwedge_{i=1}^n
(t_{\mathbf{x}} < t_{entry}^i \: \text{ or }  \:   t_{\mathbf{x}} > t_{exit}^i )\right],
 \\
      (0, 0)                    & \text{otherwise}.
  \end{cases}
  \label{eq:clipped-ray-sampling}
\end{equation}

Then we aggregate the pixels into image $\mathbf{I}_{inv}$, as the rendered image from all 3D points outside the bounding boxes. We also normally render an image from the frozen copied model, with the same camera positions, to get the reference image $\hat{\mathbf{I}}$. After that, we can calculate the reconstruction loss $\mathcal{L}_{\text{rec}}$, which defined as:
\begin{equation}
    \mathcal{L}_{\text{rec}} =  |\mathbf{I}_{inv} - \hat{\mathbf{I}} |.
\end{equation}

Finally, the overall loss to update the NeRF model for $n$ objects is formulated as:
\begin{equation}
    \mathcal{L} = \sum_{i=1}^{n}   \mathcal{L}_{\text{SDS}}^i + \alpha \mathcal{L}_{\text{rec}},
\label{total_loss}
\end{equation}
where $\alpha$ is a hyperparameter used to weigh two losses. The overall illustration of clipped ray sampling is shown in Fig.~\ref{fig:CRS}.  
We find using such scene preservation with reconstruction loss greatly helps in preventing the noise in generation and persevering the scene in placement. The effects are demonstrated in Fig.~\ref{fig:ablation} and~\ref{fig:progressive}, along with ablation studies in Sec.~\ref{sec:Ablation}.

\begin{figure}[t]
    \centering
    \vspace{-0.2cm}
    \includegraphics[width=0.7\linewidth]{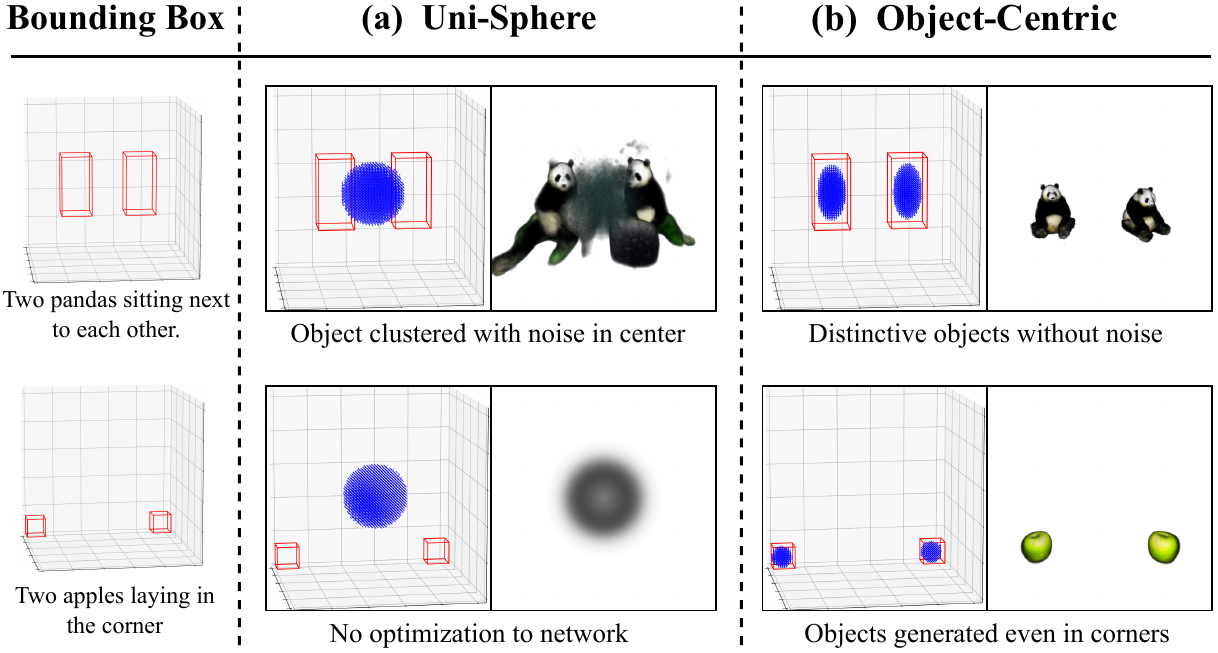}
    \vspace{-0.3cm}
    \caption{We show two toy examples in illustration of the occupancy grids with clipped ray sampling. 
    With default uni-sphere density bias (a), the objects are either clustered to the center (top), or totally missing due to gradient vanishing (bottom), while our object-centric bias (b) aligns the object's initial density with the given bounding boxes.}
    \vspace{-0.5cm}
    \label{fig:blob}
\end{figure}
\paragraph{Camera Pose Sampling.}
In scenarios with small target bounding boxes, the SDS loss with clipped ray sampling tends to be disproportionately influenced by larger objects, which contain more sampled points under constant camera distances. Such sample point discrepancy would lead to suboptimal representation or even the vanishment of smaller ones. To address this, we update the object-centric camera sampling~\cite{po2023compositional} and keep the camera distance proportional to the longest side of the bounding boxes, pulling the camera closer to smaller objects. More comprehensive technical details can be found in the \textcolor{blue}{Supplementary Material}.

\subsection{Object-Centric Density Bias Initialization}
\label{sec:object-centric init}
While clipped ray-sampling is a great way to separate the volume rendering of different objects, directly applying such a method to 3D generation frameworks, such as DreamFusion~\cite{poole2022dreamfusion}, would normally lead to poor outcomes due to the initial uni-sphere density bias in Eq.~\ref{eq:blobinit}.
As shown in the top figure of Fig.~\ref{fig:blob} (a), the objects would cluster into the center, and any density outside the bounding box would remain as noise in the final outcome. Even worse, if an object's bounding box is far from the center and has no intersection with the initial density uni-sphere, it leads to gradient vanishing and results in no optimization to objects. 
This occurrence can be attributed to the combination of Eq.~\ref{eq:clipped-ray-sampling} with the occupancy grid in Eq.~\ref{eq:grid}, wherein the condition of feed-forward becomes  $(t_{\mathbf{x}} >t_{entry} \: \text{ and }  \:   t_{\mathbf{x}} < t_{exit}  \:  \text{ and }  \:G(\mathbf{x}) = 1)$. With the occupancy grid being created from initialized density, this would entail no 3D point $\mathbf{x}$ being feed-forwarded into the MLP, yielding no optimization to the network. As shown in Fig.~\ref{fig:blob} (a) bottom, with uni-sphere density initialization, no objects are rendered and the network stays in its initial state.

To this end, we introduce object-centric density bias to accurately position the initial density within the designated bounding boxes. We adapt and modify the initialization method described in \cite{lin2023magic3d} to incorporate an object-centric approach.
More formally, for the $n$ objects with bounding boxes, the initial density bias is formulated as:
\begin{equation}
\begin{split}
      \sigma_{i}(\mathbf{x})=\lambda_\sigma \cdot\left(1-\frac{\|\frac{\mathbf{x} - c_i}{l_{i}} \|_2}{s_{\sigma}}\right),
\\
       \sigma_{\text {init }}(\mathbf{x}) = max(\sigma_{1}(\mathbf{x}), \sigma_{2}(\mathbf{x}), ..., \sigma_{n}(\mathbf{x})),
    \end{split}
\end{equation}

where $c_i$ and $l_{i}$ are the center and length of sides of the $i$-th bounding box, and $i \in \{1, 2, \ldots, n\}$. The overall density bias would be the max of all bias from the bounding boxes. 
In this way, the initialized density would be within the bounding box, with enough points satisfying the condition in both Eq.~\ref{eq:grid} and Eq.~\ref{eq:clipped-ray-sampling} while leaving minimum density occupancy outside the boundary to prevent noise. The density will also be centered in the boxes, avoiding clustering towards the scene center.
As shown in Fig.~\ref{fig:blob} (b), the initialized density is aligned with the bounding boxes, without center-clustering (top) or gradient vanishing (bottom). Moreover, the initialized density blobs are no longer strict spheres but are ellipsoids in regards to the shape of the bounding box, facilitating shape alignment.

\subsection{Generating 3D Bounding Boxes with Prompts}

\label{sec:LLM}
The bounding boxes offer a manageable way to direct the generation process. However, configuring these boxes still necessitates some manual labor, belittling the end-to-end intent of our pipeline.
As the Large Language Models (LLM) excel in understanding complex logics~\cite{lian2023llm}, we adopt a similar approach in LLM-guided diffusion \cite{lian2023llm} to use the LLM as a layout generator, creating 3D bounding boxes for a given text prompt. We follow LLM-guided diffusion in the prompting and in-context learning of the LLM as a layout generator. We make the following adjustments in the context of 3D space:

\textbf{3D Coordinate System}. We instruct the LLM (\eg GPT-3.5/4) to generate in accordance with \cite{threestudio2023} in space of [$512, 512, 512$], and provide layout examples for in-context learning.

\textbf{Containment Relationship.} Our findings suggest that conceptualizing the containment relationship (the inside/outside dynamic) as a single object could lead to improved results. Consequently, we prompt the Language Model to combine such relationships.

An example of responses can be shown below:

\noindent\textit{\underline{{Caption}}:  A pair of brown shoes placed neatly next to a black briefcase with a blue tie draped over it.}

\noindent\textit{\underline{{Objects}}: [(`a pair of brown shoes', [0, 0, 0, 256, 256, 200]),
 (`a black briefcase with a blue tie draped over it', [256, 0, 0, 256, 256, 300])]
}

The objects are represented as tuples of a description utilized in SDS loss calculation and the bounding box specified by  \textit{[x, y, z, depth, width, height]}, inspired by~\cite{lian2023llm}. This output can then directly serve as bounding boxes to guide the controlled 3D generation with methods discussed in Sec.~\ref{sec:clipped-ray-sampling} and ~\ref{sec:object-centric init}. We show the complete prompt and more textual and visualizations of the response in the \textcolor{blue}{Supplementary Material}.

\begin{figure*}[t]
    \centering
    \includegraphics[width=\linewidth]{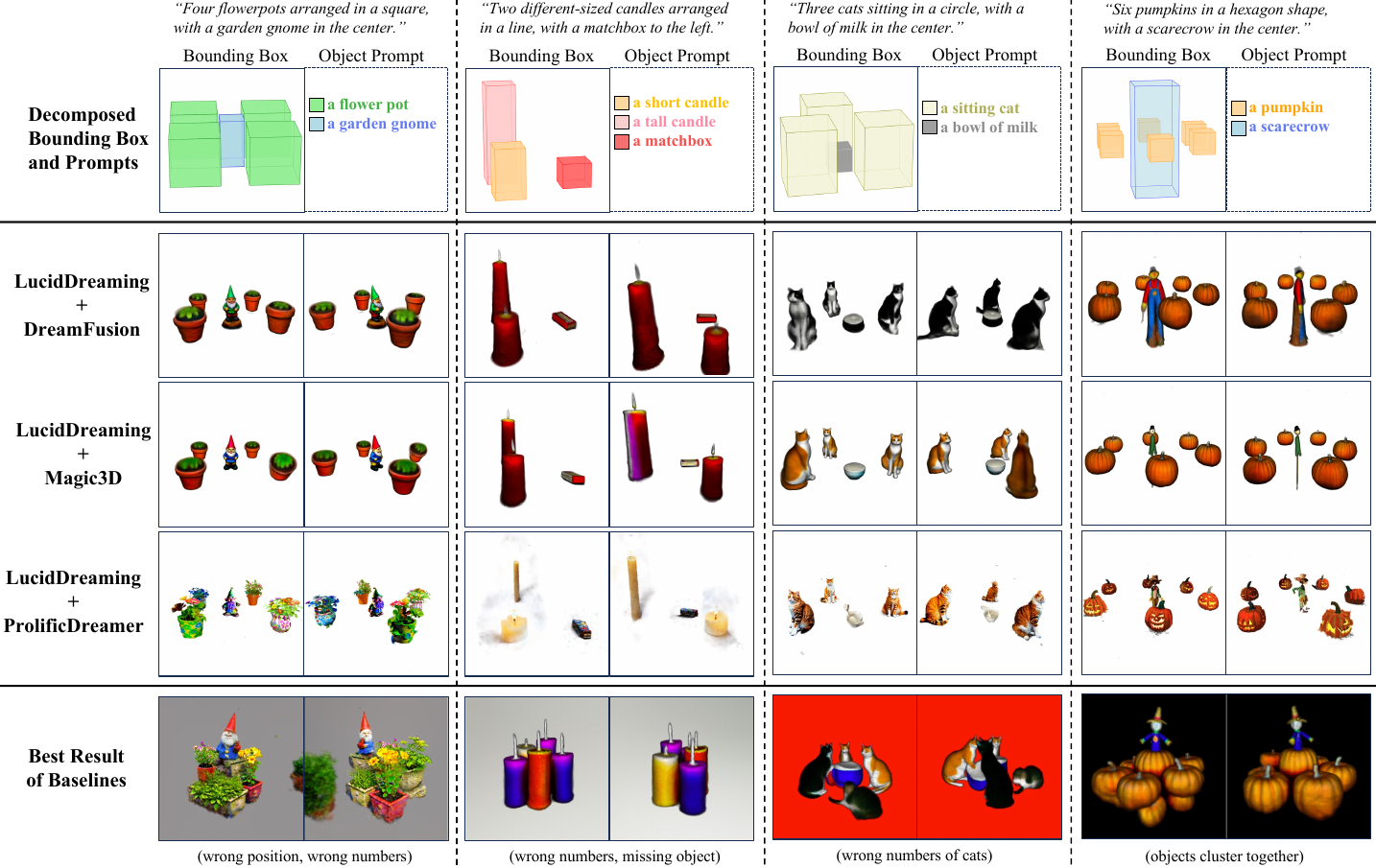}
    \caption{Examples of controlled 3D generation. The bounding boxes and prompts are decomposed from the scene prompt with an LLM. We show our method is adaptable to multiple SDS-based 3D generation methods to generate Bounding Box-controlled 3D content. In the last row, we show the best of three baseline methods with the scene prompt. Clustered objects, missing items, and wrong spatial are the most common issues in the baseline methods.  Please refer to the supplementary for more results and frameworks adapted to ours.}
    \vspace{-0.4cm}
    \label{fig:quali-gen}
\end{figure*}

\section{Experimental Results}
\label{sec:experimental_results}

This section describes the experimental results on controllable 3D generation with \ourmodel. First, we describe the implementation detail and setup in Sec.~\ref{sec:setup}. 
Then, we show the quantitative results of our method compared to baselines in Sec.~\ref{sec:Quantitative}.
In Sec.~\ref{sec:qualitative}, we present qualitative results of controlled 3D generation and placement, and Sec.~\ref{sec:Ablation} offers ablation studies on the employment of scene preservation and effects of individual components. More details are provided in the \textcolor{blue}{Supplementary Material}.

\begin{table}[t]
\begin{minipage}{0.55\textwidth}
\centering
\caption{Quantitative evaluation results with three baselines. The reported metrics are the average of individual-rated scores.}
\begin{tabular}{l|c|c|c}
  \hline
  \textbf{Model} & \textbf{BLIP-VQA} & \textbf{GPT4-V} & \textbf{Human} \\
  \hline
  DreamFusion & 20.89 & 31.08 & 41.67 \\
  \textbf{+ Ours} & \textbf{30.96} & \textbf{62.96} & \textbf{62.37} \\
  \hline
  Magic3D  & 27.08 & 42.04 & 43.70 \\
  \textbf{+ Ours} & \textbf{35.87} & \textbf{69.23} & \textbf{69.81} \\
  \hline
  ProlificDreamer & 35.09 & 41.14 & 42.60 \\
  \textbf{+ Ours} & \textbf{38.73} & \textbf{65.65} & \textbf{64.80} \\
  \hline
\end{tabular}
\label{tab:quanti}
\end{minipage}
\hfill
\begin{minipage}{0.4\textwidth}
\centering
\caption{Ablation of each component of our \ourmodel.}
    \begin{tabular}{ccc|cc}
    \toprule
    CRS  & OCDB & SP    & BLIP-VQA & GPT4-V \\
    \hline
    \xmark & \xmark       & \xmark &  27.08  & 28.13 \\
    \cmark & \xmark       & \xmark &  13.95  & 0.46     \\
    \xmark & \cmark       & \xmark &  28.86  & 43.47     \\
    \cmark & \cmark       & \xmark &  34.63  & 66.08     \\
    \cmark & \cmark       & \cmark &  35.87  & 69.23 \\
    \bottomrule
    \end{tabular}
\label{tab:ablation}
\end{minipage}
\end{table}



\subsection{Experiment Setup}
\label{sec:setup}
\paragraph{Implementation.}
In our experiments, we adopt the implementation from ThreeStudio~\cite{threestudio2023} for Text-to-3D generation frameworks and build our methods upon their codebase. We kept most of the default settings the same for generative frameworks. Notably, We adjusted all sampling resolutions for SDS to 128 and the training steps to 10,000 for fair comparisons. We also 
use Instant-NGP as NeRF representation for all methods in accordance with~\cite{threestudio2023}.

\paragraph{Baseline Models.} 
Our experiments mainly focus on implementing and comparing with three SDS-based Text-to-3D generative frameworks: DreamFusion~\cite{poole2022dreamfusion}, Magic3D~\cite{lin2023magic3d}, and Prolific Dreamer~\cite{wang2023prolificdreamer}. Following ~\cite{threestudio2023}, we use Deep Floyd IF~\cite{saharia2022photorealistic,deepFloyd2023IF} as the Distillation Model for DreamFusion and Magic3d, and Stable Diffusion 2.1~\cite{rombach2022high} for ProlificDreamer.

\paragraph{Dataset.} 
To facilitate controllable 3D generation and establish a standard benchmark for future work, we provide a dataset that consists of 150 scene prompts, each accompanied by corresponding bounding boxes and object descriptions. The prompts predominantly feature scenes with multiple objects and emphasize spatial relationships, numeracy, and various sizes of 3D objects. Specifically, we gather 3D bounding boxes generated from GPT-4~\cite{openai2023gpt4} and subsequently validate them with human annotators. We will release the dataset for future research.


  


\subsection{Quantitative Results}
\label{sec:Quantitative}

\paragraph{Evaluation Metrics.}

In this paper, we include three metrics to quantify the controllability of 3D object generation. First, we follow the T2I benchmark~\cite{huang2023t2i} for object-wise evaluation with BLIP-VQA~\cite{li2022blip}. 
The BLIP-VQA independently verifies each object in the prompt and aggregates the probabilities to a final score.
Second, we also adopt the Chain-of-Thought evaluation prompt from T2I benchmark~\cite{huang2023t2i}, which requires the model to explain the image first and then relate to the given prompt and give a rating from 0 to 100. We use such prompting with the GPT4-Vision (GPT4-V)~\cite{openai2023gpt4}\footnote{About 1\% of evaluated images inexplicably trigger the OpenAI safety protocol, so we omit those in the calculation of the average ratings.} to evaluate the results.
Eventually, we gather feedback from human annotators for the alignment between text and image. We ask the annotator to rate from 1 to 5 and normalize the options to 0-100. To avoid human biases, we gather at least three annotations per image and report the average. 

\vspace{+1mm}
We show the quantitative results in Tab.~\ref{tab:quanti}. It can be clearly seen that \ourmodel achieves significant improvements in controllability than the baseline methods.
For example, we achieve 8.8\% and 26.9\% improvements compared to Magic3D results with BLIP-VQA and GPT4-V, which demonstrates our method outperforms the baseline results with significant improvements in terms of both 3D quality and controllability. In addition, our method shows notable enhancements in human evaluations and preferences, emphasizing its superiority in controlled 3D generation with improved quality and precision.
\vspace{-0.5cm}
\paragraph{Discussion.} While CLIP scores are commonly used for evaluating 3D generation for quality measurement, their suitability for assessing controllability is limited~\cite{huang2023t2i}. We provide detailed comparative results and discussions in the \textcolor{blue}{Supplementary Material}.

\begin{figure}[t]
    \centering
    \includegraphics[width=\linewidth]{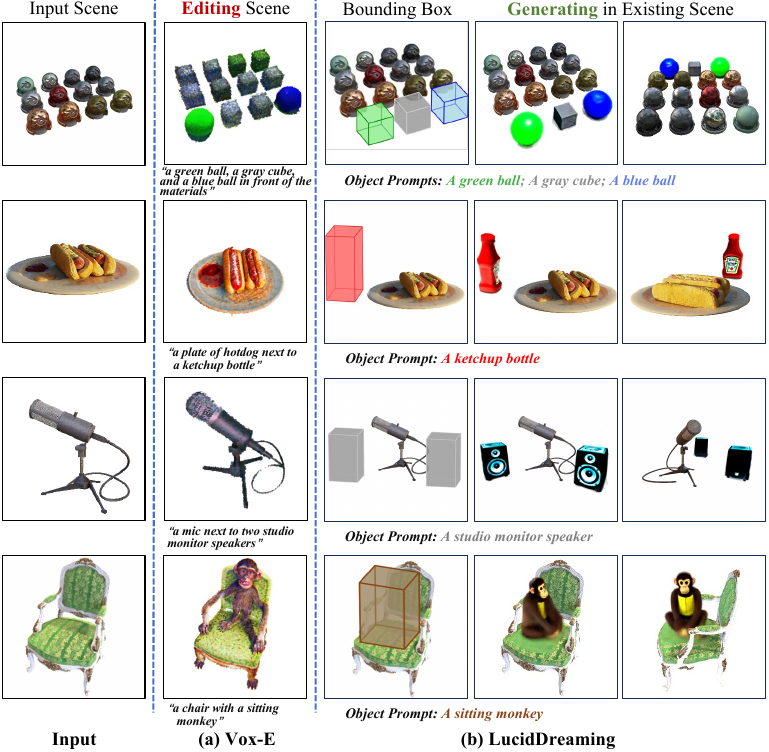}
    \vspace{-0.7cm}
    \caption{
    Unlike existing methods focused on global scene editing (\eg, Vox-E), our approach enables the controllable generation of high-quality 3D content while preserving the original scene.
    }
    \vspace{-0.3cm}
    \label{fig:quali-place}
\end{figure}

\vspace{-0.5mm}
\subsection{Qualitative Results}
 
\label{sec:qualitative}
\vspace{-0.5mm}
\paragraph{Controlled 3D Generation.}
We provide qualitative results of controlled 3D generation in Fig.~\ref{fig:teaser} (top) and Fig.~\ref{fig:quali-gen}. Results demonstrate that our method can effectively exert control over 3D generation given merely a scene prompt. More specifically, we show the scene prompt and decomposed 3D bounding boxes with object descriptions in the first row of Fig.~\ref{fig:quali-gen}, which are used as conditions for controllable object-centric 3D generation with \ourmodel. The results in the middle rows of Fig.~\ref{fig:quali-gen} and the top-right of Fig.~\ref{fig:teaser} highlight that our method is capable of generating objects closely adhered to the controlling bounding boxes and the scene prompts. Moreover, these results serve as examples demonstrating the integration of \ourmodel with various 3D generation methodologies, confirming its adaptability and compatibility across a wide range of 3D generation frameworks.

In comparison, we display the best (in terms of GPT4-V rating) among the three baseline methods in the bottom row of Fig.~\ref{fig:quali-gen}. It is observed that missing items, wrong positions, and object clustering to the center are not uncommon in baseline generations. This is largely attributed to the diffusion model's deficiency in understanding complex prompts, such as those with numeracy and spatial relationships, and the limitations of constraining only from 2D rendering, lacking explicit 3D controls. This confirms that relying exclusively on text prompts for conditioning and controlling 3D model generations leads to unreliable outcomes.

\paragraph{Controlled Object Placement to Scene.}
In Fig.~\ref{fig:teaser} (bottom) and Fig.~\ref{fig:quali-place}, we show qualitative results of content generation and placement within a pre-trained NeRF representation on the blender-synthetic dataset~\cite{mildenhall2021nerf}.
In Fig.~\ref{fig:quali-place} (right), we demonstrate our capacity to generate objects given defined bounding boxes while maintaining the overall integrity and composition of the original scene. In our experiments, we choose Vox-E~\cite{sella2023vox} as the representative editing method for the placement comparison, and generate content in blender-synthetic objects in an editing approach.
As depicted in Fig.~\ref{fig:quali-place} (middle), when conditioned solely on text prompts and lacking precise control mechanisms, Vox-E primarily alters the semantic properties of existing objects rather than creating new elements in the areas designated by the prompts. Instead, \ourmodel is able to create full 3D objects with more fine-grained control.

\vspace{-0.05cm}
\subsection{Ablation Studies}
\begin{figure}[t!]
    \centering
    \includegraphics[width=0.8\linewidth]{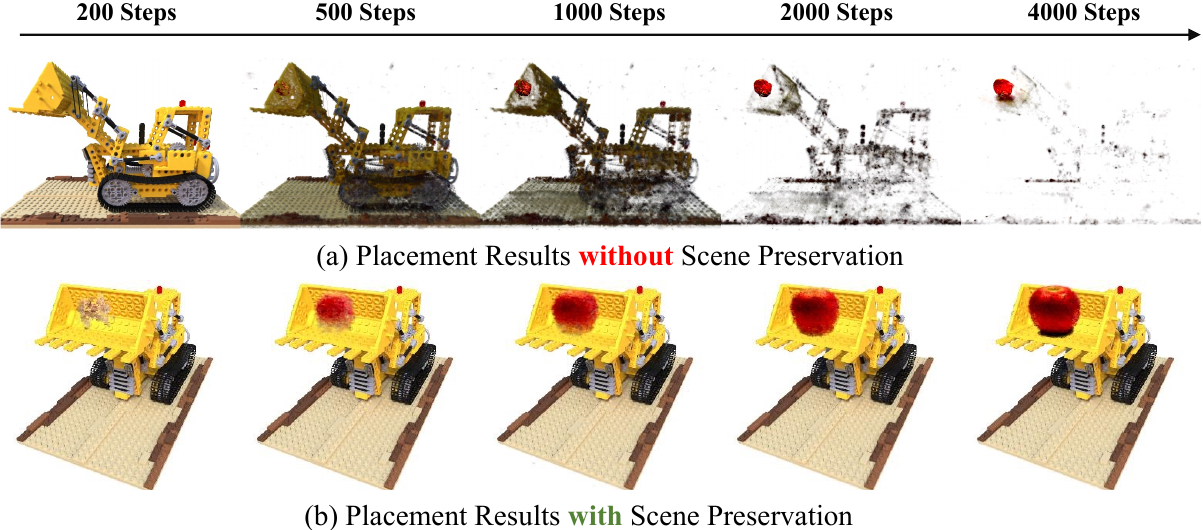}
    \vspace{-0.2cm}
    \caption{
    Ablation on scene preservation with reconstruction loss in pre-trained scenes, validating its effectiveness in preserving the integrity of the original objects.
    }
    \vspace{-0.4cm}
    \label{fig:progressive}
\end{figure}

\vspace{-0.05cm}
\label{sec:Ablation}
We find the scene preservation discussed in Sec.~\ref{sec:clipped-ray-sampling} plays an important role in the generation quality. As illustrated in Fig.~\ref{fig:ablation}, this approach effectively eliminates noise in the area outside the bounding boxes. In addition, it preserves the original object intact when placing new content in a pre-trained NeRF scene. In contrast, naively resuming from the pre-trained NeRF and generating object would lead to catastrophic forgetting and destruction of the integrity of the original object. We present a progressive comparison in Fig.~\ref{fig:progressive}, demonstrating that scene preservation effectively prevents the bulldozer from diminishing. 

\begin{wrapfigure}{r}{0.6\textwidth}
    \vspace{-30pt}
    \centering
    \includegraphics[width=\linewidth]{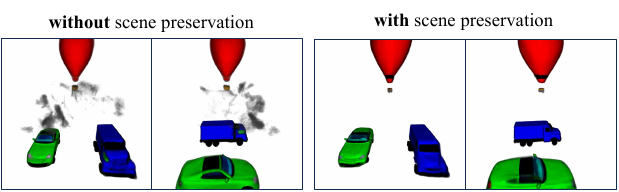}
    \vspace{-0.5cm}
    \caption{Ablation experiments on scene preservation in generation from scratch. It helps to prevent noises during the generation.}
    \label{fig:ablation}
    \vspace{-20pt}
\end{wrapfigure}

\paragraph{Ablations on \ourmodel \xspace Components}
In Tab.~\ref{tab:ablation}, we provide ablations studies on components in \ourmodel with Magic3D as the backbone generation framework, including the proposed Clipped Ray Sampling (CRS), Object-Centric Density Bias (OCDB), and Scene Preservation (SP). Specifically, with only CRS, \ourmodel fails hard with unsatisfying BLIP-VQA and GPT4-V rating, primarily due to gradient vanishing discussed in Fig.~\ref{fig:blob}.
Using OCDB slightly improves the controlled generation, but it achieves a much higher rating when combined with CRS and constrained by bounding boxes. The SP further prevents floaters and noises and enhances the final quality.

\vspace{-0.15cm}
\section{Conclusion}
\vspace{-0.15cm}
In this paper, we demonstrate that the existing Text-to-3D generation methods lack fine-grained controllability, while controllable counterparts are limited due to their reliance on customized diffusion models and their inability to easily adapt to other pipelines.
To overcome these challenges, we introduce LucidDreaming to achieve controllable object-centric 3D generation from scratch or pre-trained scenes. We propose clipped ray sampling and object-centric density bias initialization to generate discrete 3D objects that conform to given specifications.
Our method can enhance the controllability of various 3D generation techniques in a plug-and-play manner, achieving fine-grained control over the created contents. We include more discussions on limitations and future works in the \textcolor{blue}{Supplementary Material}.

\clearpage
\newpage
\appendix

\section{Overview of Supplementary Material}
The supplementary material is organized into the following sections:

\begin{itemize}
    \item Section~\ref{tech_details}: Technical Details of \ourmodel.

    \item Section~\ref{data_detail}: Dataset and Evaluation Details.
    \eric{\item Section~\ref{application}: We show more applications of our method.}

    \item Section~\ref{visualization}: More visualizations on four different adapted pipelines, including DreamFusion~\cite{poole2022dreamfusion}, Magic3D~\cite{lin2023magic3d}, ProlificDreamer~\cite{wang2023prolificdreamer} and Zero123~\cite{liu2023zero}.

    \item Section~\ref{limitation}: Broader Impact and Limitation.
\end{itemize}

\section{Technical Details}
\label{tech_details}
\subsection{Camera Pose Sampling}
To promote object distinctiveness and generation quality, we apply object-centric camera pose sampling to make sure all the objects are positioned in the center of the rendered view from the camera for SDS, inspired by \cite{po2023compositional}. Specifically, for the $i$-th object with its bounding box, where $i \in \{1, 2, \ldots, n\}$ as $n$ objects in the scene, we modify the original sampled ray from the camera  $\mathbf{r}(t)=\mathbf{o}+t \mathbf{d}$ to point to the object's bounding box, with a directional offset $d_\text{center} = c_i - c_\text{scene}$, where $c_i$ and $c_\text{scene}$ represent the center of the $i$-th bounding box and the overall scene. 

Additionally, we also observed biases existing if the sizes of boxes are disproportional in the scene and the larger boxes, with more sampled points under constant camera distance, would dominate the optimization and lead to poor quality or even vanishment of the smaller ones. So we adjust the camera distance to be proportional to the longest side of the object bounding box, with another offset $d_\text{scale} = (c_i - \mathbf{p}) \times \frac{(max(l_\text{scene}) - max(l_i))}{\beta \times (max(l_\text{scene})}$ , in which $\mathbf{p}$ is the principle point of current camera. $l_\text{scene}$ and $l_i$
 are the sides of scene and $i$-th bounding box, respectfully. $\beta$ is a hyperparameter to weigh the ratio of camera distance and box size. 

In summary, for the $i$-th object with its bounding box, the updated object-centric camera sampling leads to the sampled ray as:
\begin{equation}
\resizebox{.60\hsize}{!}{$
\begin{aligned}
    &{\mathbf{r_i}(t)} = d_\text{center} + d_\text{scale} + \mathbf{o} + t \mathbf{d}, \\
    &{d_\text{center}} = c_i - c_\text{scene}, \\
    &{d_\text{scale}} = (c_i - \mathbf{p}) \times \frac{\left(\max(l_\text{scene}) - \max(l_i)\right)}{\beta \times \max(l_\text{scene})}.
\end{aligned}
\label{eq:oc-cam-sample}
$}
\end{equation}

We find such camera pose sampling empirically improves the generation quality of individual objects combined with clipped ray sampling.

\subsection{Implementation Details}

We follow the implementation from ThreeStudio~\cite{threestudio2023} and keep most of the hyper-parameters the same. Particularly, we changed the sample resolution to 128 and training steps to 10,000 for all backbone methods for fair comparison. We use a $\alpha$ of 0.3 in Eq.~\ref{total_loss} of the manuscript, and a $\beta$ of 1.0 in camera pose sampling above for all experiments.

\vspace{-0.2cm}
\section{Evaluation Details and Dataset}
\vspace{-0.1cm}

\label{data_detail}
\subsection{Dataset Construction}
\vspace{-0.1cm}

\begin{figure}[h!]
    \centering
    \begin{subfigure}[b]{0.48\textwidth}
        \centering
        \includegraphics[width=\linewidth]{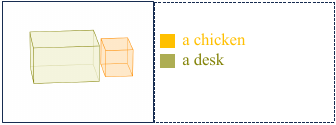}
        \caption{Illustration of normal example}
        \label{fig:normal_example}
    \end{subfigure}
    \hfill
    \begin{subfigure}[b]{0.48\textwidth}
        \centering
        \includegraphics[width=\linewidth]{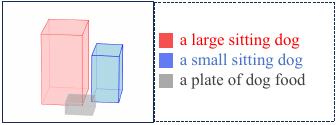}
        \caption{Illustration of complex example}
        \label{fig:complex_example}
    \end{subfigure}
    \caption{Examples of bounding boxes in the dataset.}
    \label{fig:bb_examples}
\end{figure}

We constructed our dataset with the help of GPT-4 for prompting. The prompts and bounding boxes are generated from GPT-4~\cite{openai2023gpt4} and verified by human annotators to ensure correctness and diversity. More specifically, we ask human annotators to abandon examples with mismatched bounding boxes and repetitive 3D objects. We use a temperature of 0.5 in the GPT4 prompting. More examples can be found in the Fig.~\ref{fig:quali-gen} in the manuscript. We include two subsets in the dataset, normal and complex, which contain 50 and 100 prompts respectively. We provide the details below and in Fig.~\ref{fig:bb_examples}:

\vspace{-0.4cm}
\paragraph{Normal.} In the normal subset, there are two different objects with one simple spatial relationship, such as ``next to", or ``on the right of". For instance:

\noindent\textit{\underline{{Caption}}:  a chicken near a desk.}

\noindent\textit{\underline{{Objects}}:  [(`a desk', [156, 106, 200, 200, 300, 150]),
 (`a chicken', [156, 436, 200, 150, 76, 112])]
}

\vspace{-0.2cm}

\paragraph{Complex.} We also provide a larger subset for more complex prompts, containing objects in numbers ranging from 2 to 7. There are also numbers and implications in the prompt, such as ``Four different-sized \textbf{xxx}". One example is shown:

\noindent\textit{\underline{{Caption}}: Two dogs sitting side by side, one larger than the other, with a plate of dog food in front.}

\noindent\textit{\underline{{Objects}}:   [(`a large sitting dog', [156, 106, 0, 200, 150, 300]),
 (`a small sitting dog', [156, 256, 0, 150, 100, 200]),
 (`a plate of dog food', [356, 206, 0, 100, 100, 50])]}

\begin{figure*}[t]
   \centering
    \includegraphics[width=0.9\linewidth]{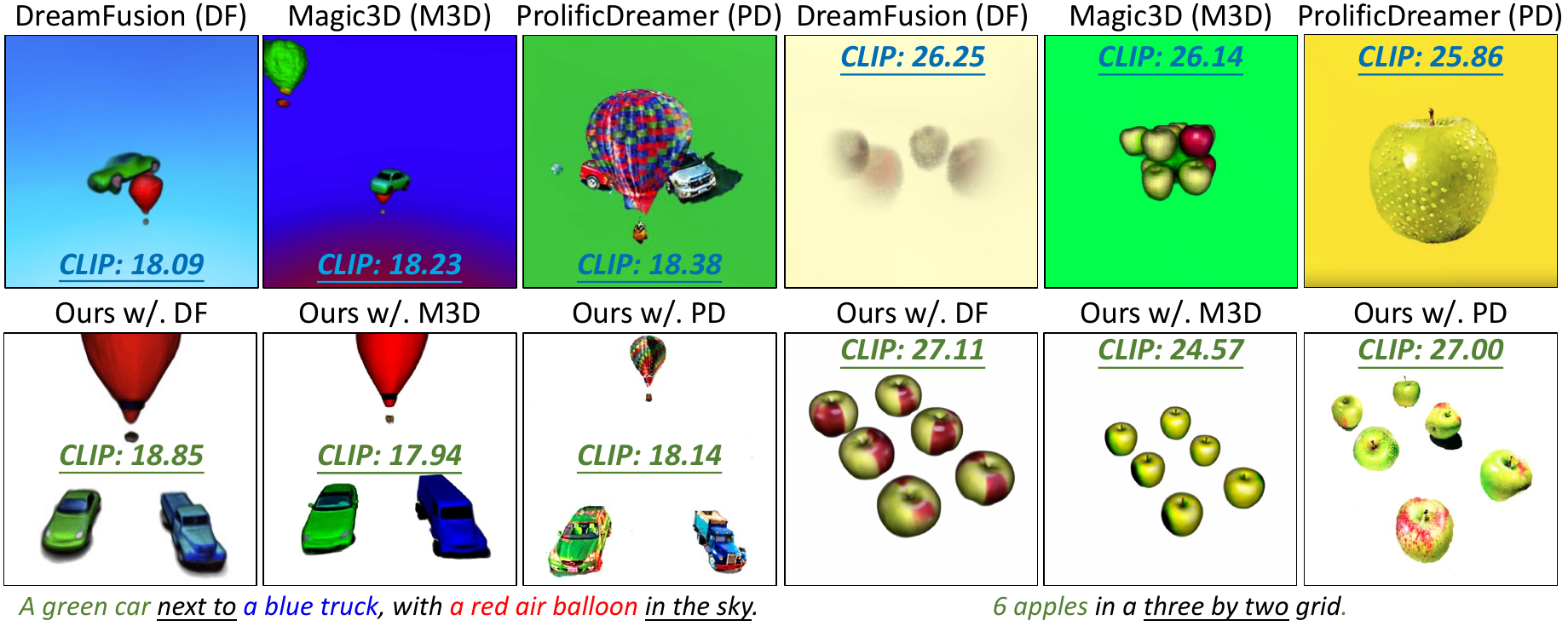}
    \caption{We show 2 examples to illustrate that CLIP score is not suitable for the goal of controllability because it only evaluates whether objects appear in the picture, but ignores the number of objects or spatial position relationships. For example, using the Magic3D as backbone framework, our method generates a scene clearly following the prompt while the base Magic3D only has clustered apples. It is obvious that the CLIP score fails to capture the essence of the prompt and thus gives unreliable ratings for controllability.}
    \label{fig:clip}
\end{figure*}

The bounding boxes are specified as \textit{[x, y, z, depth, width, height]}, in the 3D space of [512, 512, 512]. The 3D coordinate system follows the one in ThreeStudio~\cite{threestudio2023}.
\subsection{Evaluation}
\subsubsection{On CLIP Assessment of Controllability in 3D Generation}

In recent years, it has become a standard procedure to evaluate 3D generative frameworks with CLIP~\cite{radford2021learning} encoders. Although the CLIP score reflects to a certain extent whether the generated 3D object matches the input text prompt, this indicator is still far from a reasonable assessment of controllability. Following the previous works~\cite{po2023compositional, poole2022dreamfusion}, we show the overall CLIP score and CLIP R-Precision of all methods in Tab.~\ref{tab:clip_score_comparison}. Though \ourmodel achieves dominating visually-better results, the CLIP score is highly close to our baseline results.

As many previous works point out, diffusion models, using text encoders for prompt comprehension, inherently have trouble understanding hard logics~\cite{lian2023llm, zhang2023controllable}. CLIP, as encoders itself, also exhibits limitations in capturing and interpreting complex logical relationships or spatial details~\cite{huang2023t2i}. Thus, it is counter-intuitive to use CLIP as a reliable metric to measure controllability.
As illustrated in Fig.~\ref{fig:clip},
although the generated 3D scene does not reasonably reflect the description of the objects in the text prompt, such as the number of objects, and spatial position relationship, the CLIP scores of the baseline methods are still comparable to ours, which clearly shows that the CLIP score is not suitable for controllability goal.

\begin{table}[h]
\vspace{-0.5cm}

\centering
\begin{tabular}{|c|c|c|}
\hline
\textbf{Method} & \textbf{CLIP Score} & \textbf{CLIP R-Precision} \\
\hline
Ours & 28.13 & 51.94\\
Baselines & 28.12 & 50.11\\
\hline
\end{tabular}
\vspace{+0.3cm}
\caption{Comparison of CLIP Scores and CLIP R-Precision of all methods (DreamFusion + Magic3D + ProlificDreamer), we use the clip-vit-base-patch16 for CLIP encoders and prompts from \textit{spatial} subset in T2I Benchmark~\cite{huang2023t2i} for R-Precision evaluation. }
\vspace{-1cm}
\label{tab:clip_score_comparison}
\end{table}

\subsubsection{GPT4-V Evaluation}

\paragraph{Prompting GPT4-V with Chain-of-Thought} 
Following~\cite{huang2023t2i}, we use a prompt that encourages Chain-of-Thoughts for GPT4-V~\cite{openai2023gpt4} to evaluate the images with corresponding scene descriptions. Specifically, the GPT4-V would first explain the images and then rate their alignment with the prompts with a score ranging from 0 to 100, the higher the better. The full prompt and API request are given in Tab.~\ref{tab:gpt4-v-prompt}. Such prompt is utilized for all evaluations across the methods, and we use a max token of 2,000 with a temperature of 0.5 for the GPT-4 API parameters.

\begin{figure*}[t!]
   \centering
    \includegraphics[width=\linewidth]{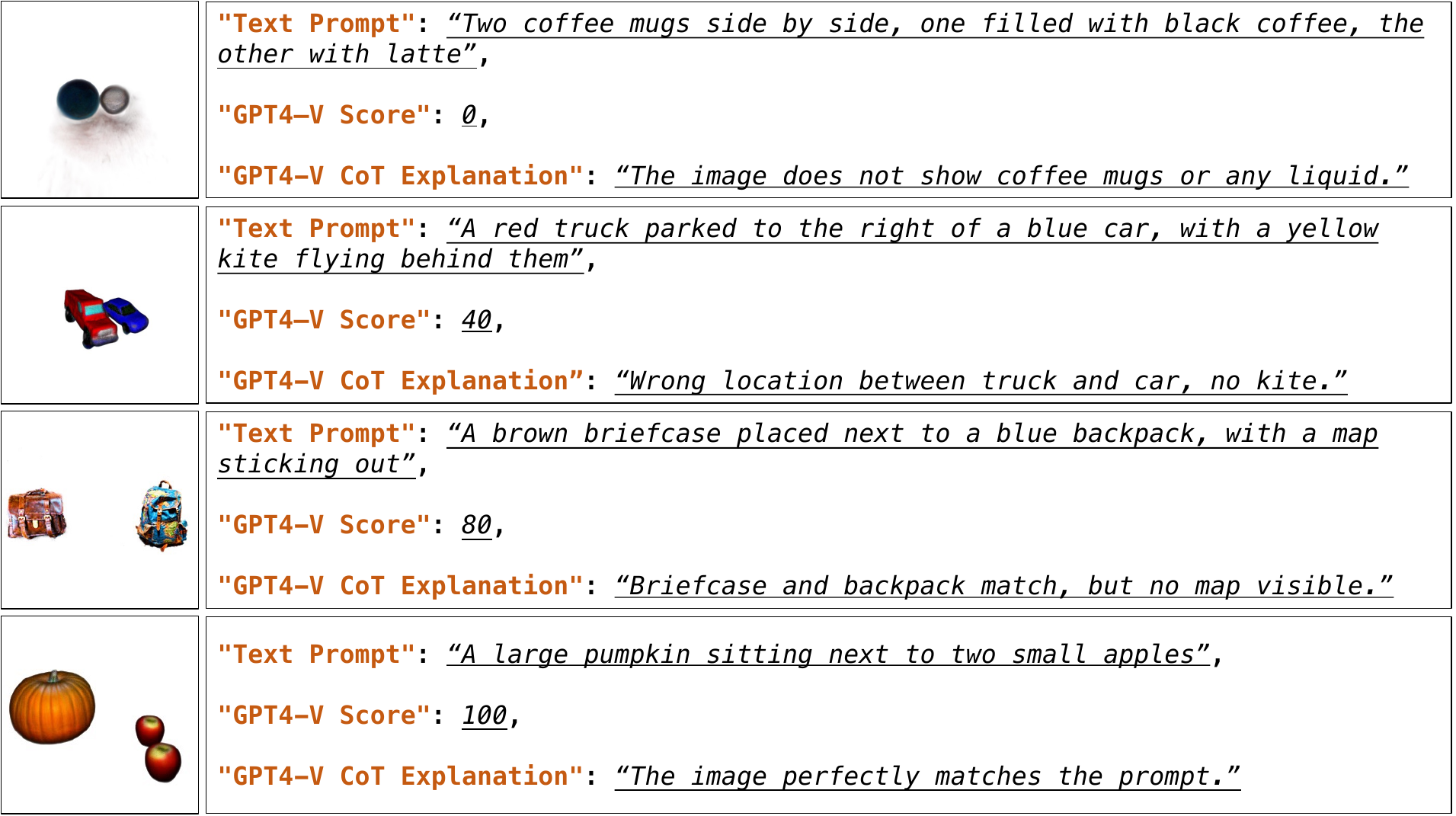}
    \vspace{-0.4cm}
    \caption{Examples of GPT4-V evaluation. The left column is the input image, and the right column is the text prompt and corresponding responses from GPT4-V API. It can be seen that GPT4-V returns a reasonable score and the chain-of-thought (CoT) explanation.}
    \label{fig:gpt4-v}
\end{figure*}

\begin{figure*}[t!]
   \centering
    \includegraphics[width=\linewidth]{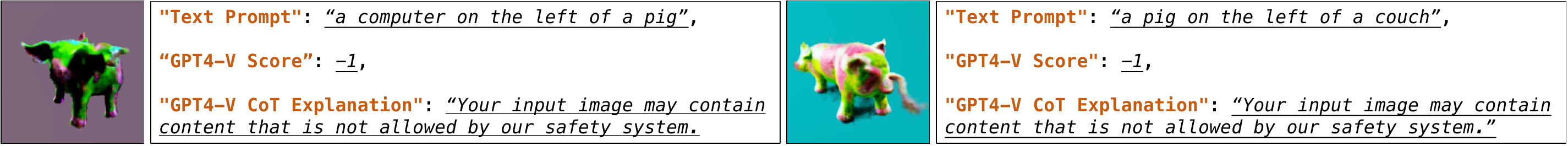}
    \vspace{-0.5cm}
    \caption{A small number of samples unexpectedly trigger the security mechanism of GPT4-V, and we ignore these images when calculating the final average score.}
    \label{fig:safty_triggers}
\end{figure*}

We show examples of image and GPT4-V response in Fig.~\ref{fig:gpt4-v}. It can be clearly seen that GPT4-V can successfully explain the image, align with the prompt, and evaluate the image. As mentioned in the main manuscript, a small amount of images triggers the safety protocol of OpenAI and they decline the requests. Thus, we ignore these images in evaluating the overall average. We also show such samples in Fig.~\ref{fig:safty_triggers}.

\subsection{Human Annotations}
\begin{figure*}[!]
   \centering
   \vspace{5.0cm}
    \includegraphics[width=1.0\linewidth]{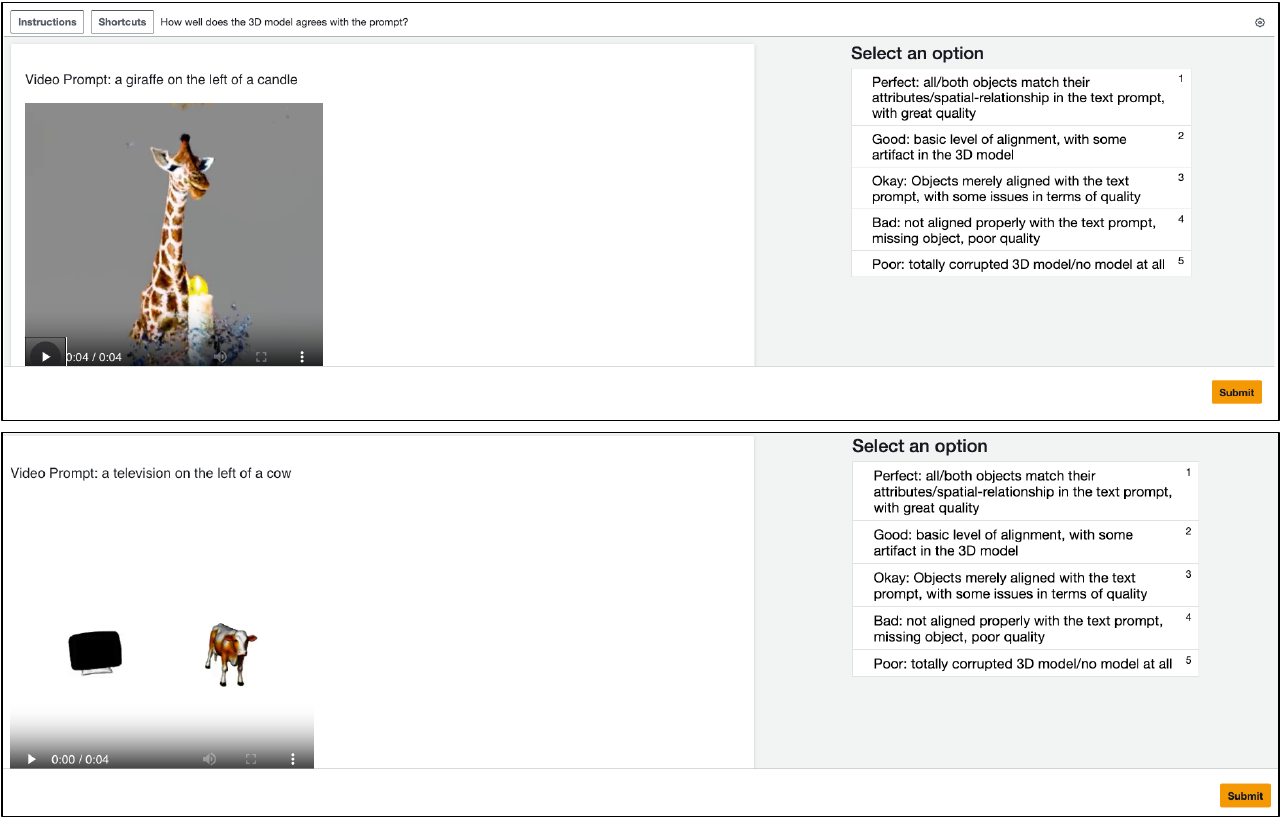}
    \caption{Visualization of Amazon Mechanical Turk (AMT) for Human Annotations.}
    \label{fig:vis_amt}
\end{figure*}

Human evaluation experiments were carried out using Amazon Mechanical Turk (AMT). In these experiments, annotators were tasked with assessing the degree of match between the images generated and the corresponding text prompts.  Each pair was rated on a scale of 1 to 5 by three separate human annotators, based on how well the image aligned with the text. This process is demonstrated in Fig.~\ref{fig:vis_amt}, which display the examples of evaluation interfaces. To alleviate human bias, we make sure each image-prompt pair receives at least three ratings from annotators and take the average. 

\subsection{More ablation studies}
\paragraph{Ablations on Scene Preservation.}
We find the scene preservation discussed in Sec.~\ref{sec:clipped-ray-sampling} plays an important role in the generation quality. As illustrated in Fig.~\ref{fig:ablation}, this approach effectively eliminates noise in the area outside the bounding boxes. In addition, it preserves the original object intact when placing new content in a pre-trained NeRF scene. In contrast, naively resuming from the pre-trained NeRF and generating object would lead to catastrophic forgetting and destruction of the integrity of the original object. We present a progressive comparison in Fig.~\ref{fig:progressive}, demonstrating that scene preservation effectively prevents the bulldozer from diminishing. 
\begin{figure}[t!]
    \centering
    
    \includegraphics[width=0.8\linewidth]{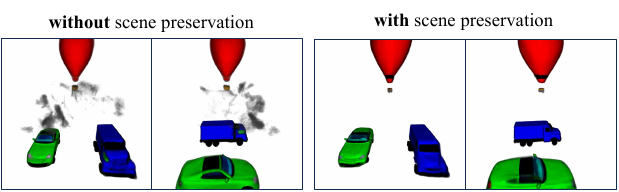}
    \vspace{-0.3cm}
    \caption{
    Ablation experiments on scene preservation in generation from scratch. It helps to prevent noises during the generation.}
    \vspace{-0.1cm}
    
    \label{fig:ablation}
\end{figure}

\section{More Applications}
\label{application}
\begin{figure}[t]
    \centering
    \includegraphics[width=\linewidth]{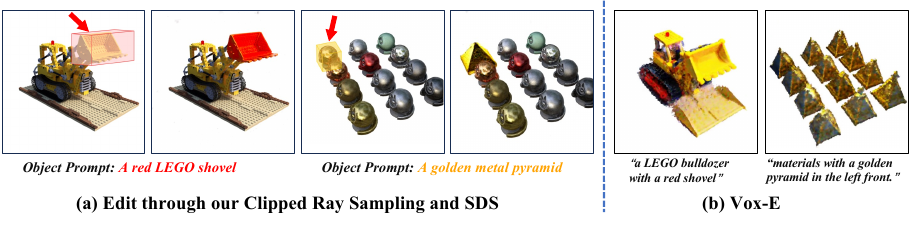}
    \vspace{-0.7cm}
    \caption{
Given a bounding box from the user, our Clipped Ray
Sampling can also achieve editing with SDS with precise control.
Also, existing 3D editing methods cannot achieve local editing.
    }
    \label{fig:quali-edit}
\end{figure}

\eric{
In addition to placing objects into the new scene, we also demonstrate the 3D editing and interaction with original scene components as a use case of Clipped Ray Sampling. In Fig.~\ref{fig:quali-edit}, we show examples that achieve precise editing on the desired location, indicated by the bounding box provided by the user. In contrast, without such capabilities and using only text prompts as the venue of controlling conditions, previous work~\cite{sella2023vox} fails to accurately alter the scene as desired. Our method is able to alter the existing NeRF scene within the bounding boxes while keeping the parts outside intact.
}

\section{More Visualizations}
\label{visualization}
\begin{figure*}[!]
   \centering
    \includegraphics[width=\linewidth]{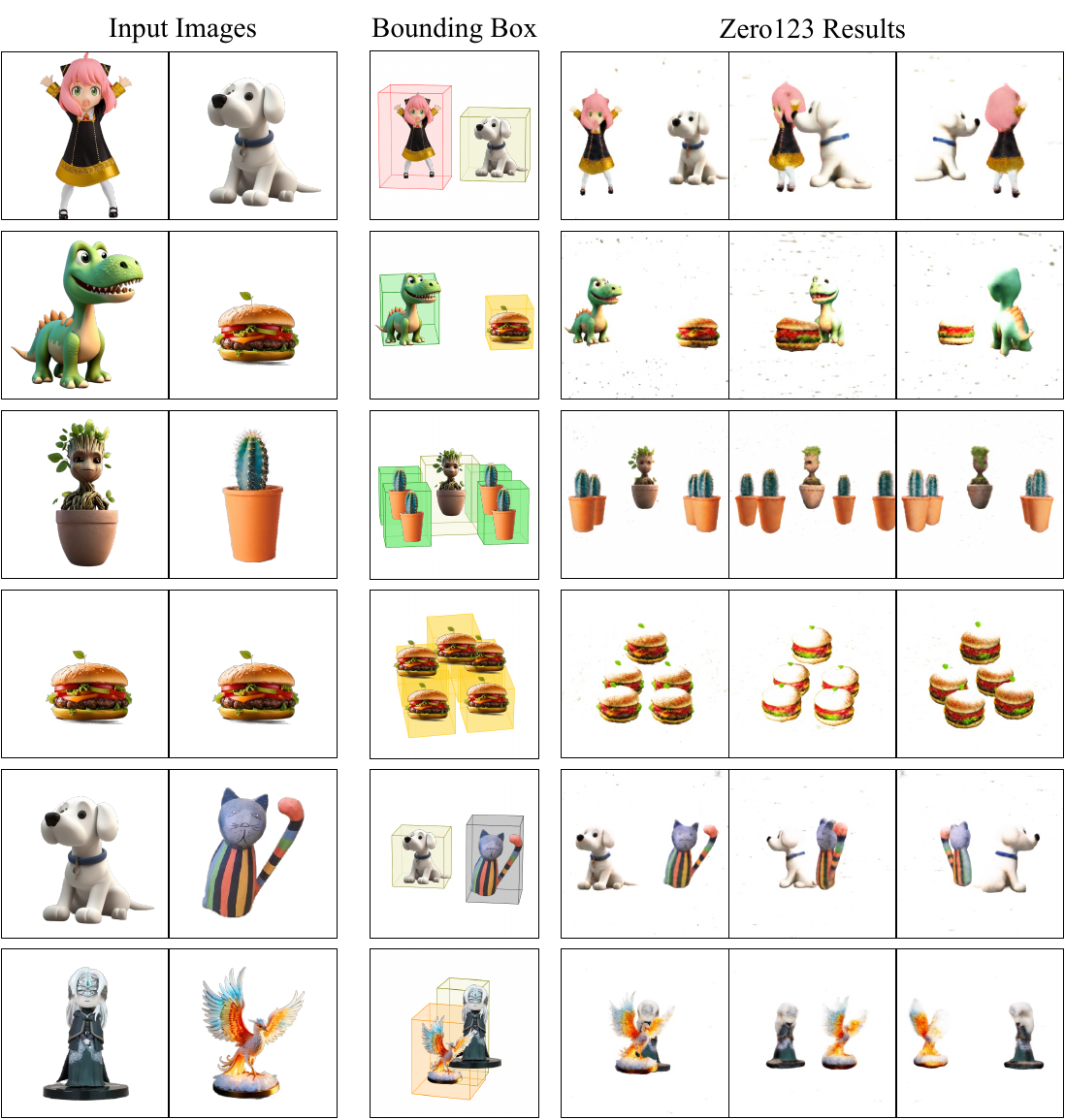}
    \caption{More visualization results of our \ourmodel with \textbf{Zero123}~\cite{liu2023zero} backbone.}
    \label{fig:more_vis_z123}
\end{figure*}

In addition to three baselines as mentioned in the main paper, we also show the visual results of our \ourmodel on Zero123~\cite{liu2023zero}, which is an image-to-3D generation pipeline. As illustrated in Fig.~\ref{fig:more_vis_z123}, the generated 3D objects are well aligned with the information of the input image, strictly following the bounding box given for each object in the scene.

\begin{figure*}[!]
   \centering
    \includegraphics[width=1\linewidth]{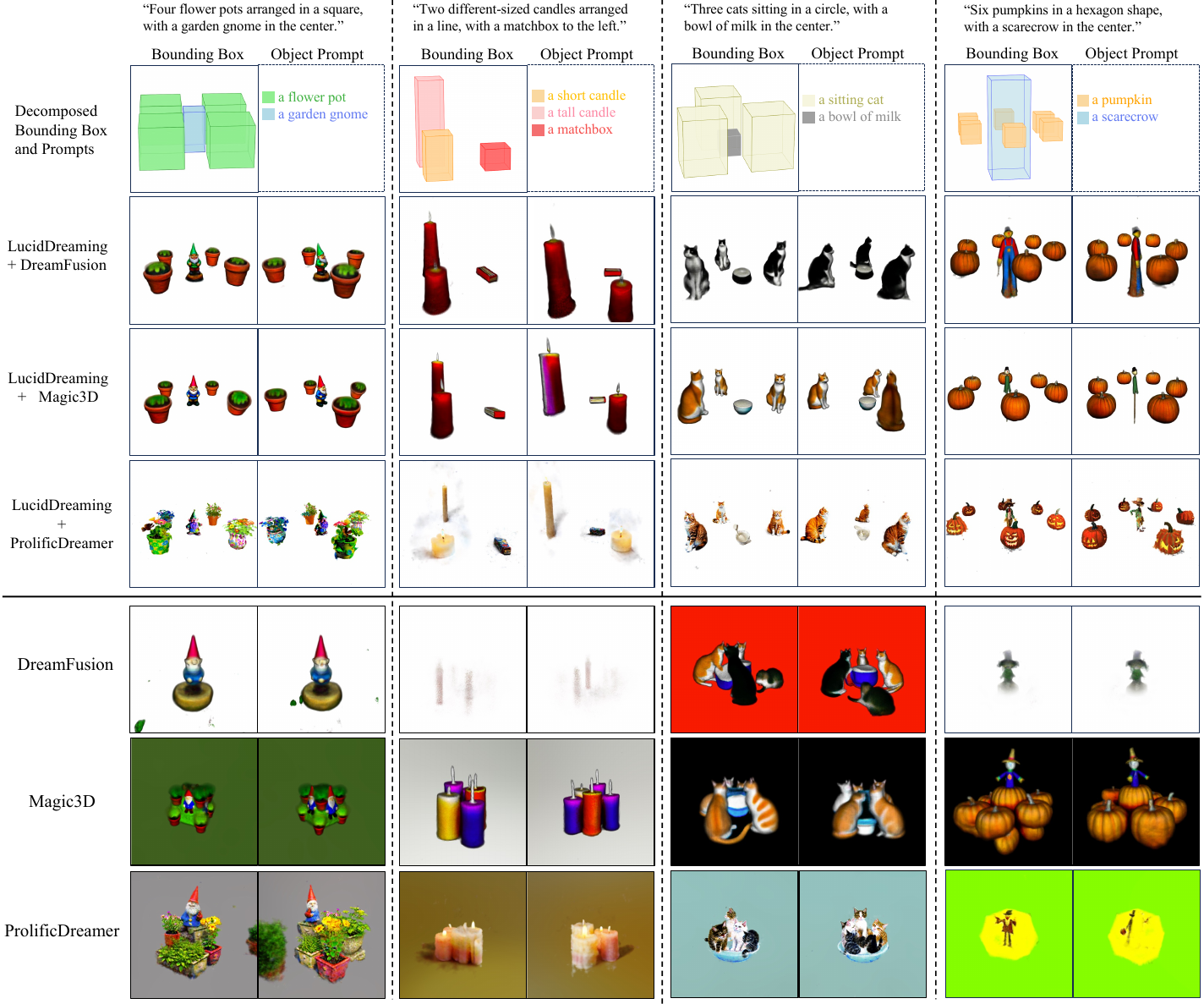}
    \caption{More visualization results of our \ourmodel compared to baseline methods. We show the 3D bounding box, text prompt, and two different views.}
    \label{fig:vis_1}
\end{figure*}

\begin{figure*}[!]
   \centering
    \includegraphics[width=\linewidth]{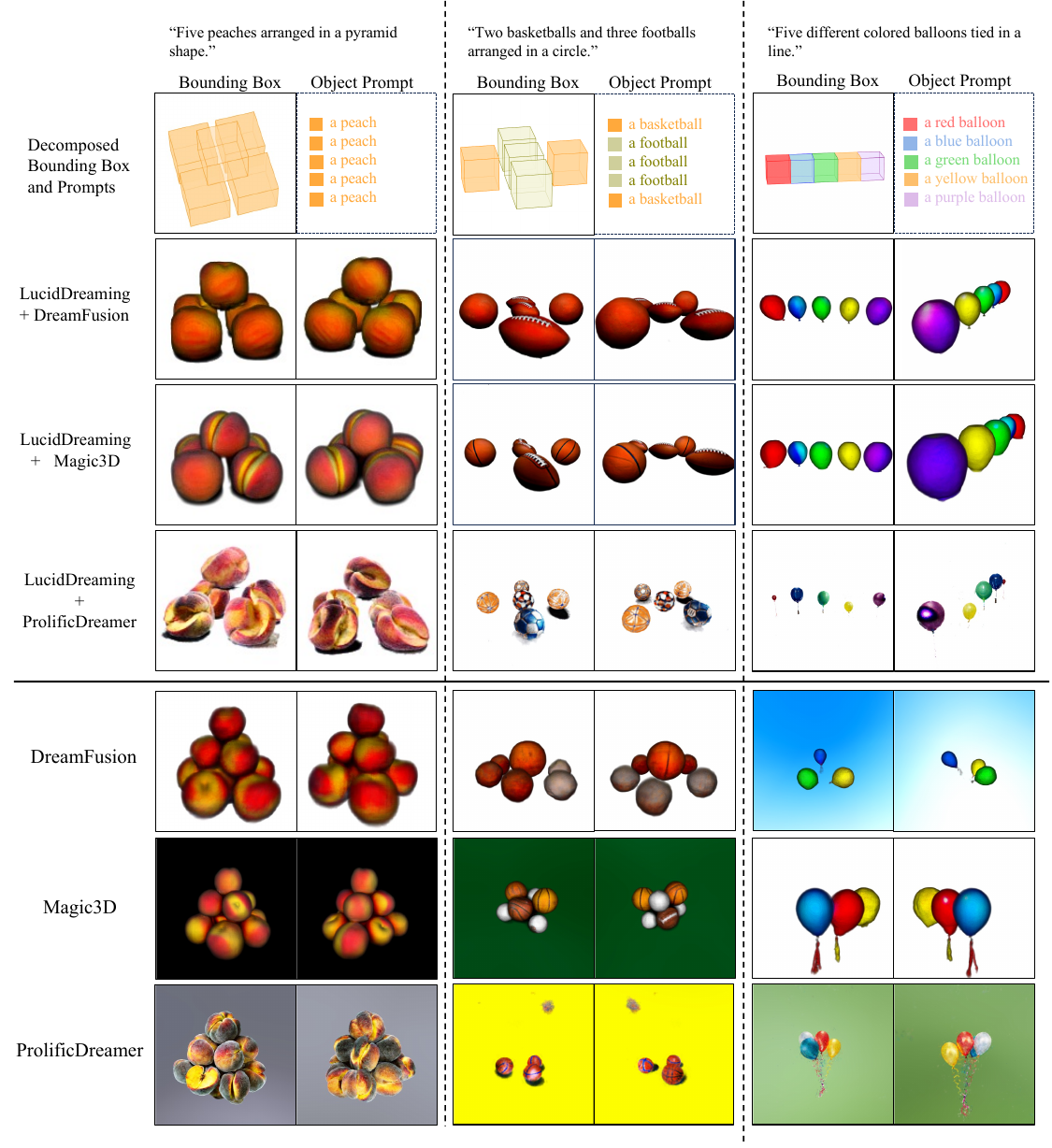}
    \caption{More visualization results of our \ourmodel compared to baseline methods. We show the 3D bounding box, text prompt, and two different views.}
    \label{fig:vis_2}
\end{figure*}

\begin{figure*}[!]
   \centering
    \includegraphics[width=\linewidth]{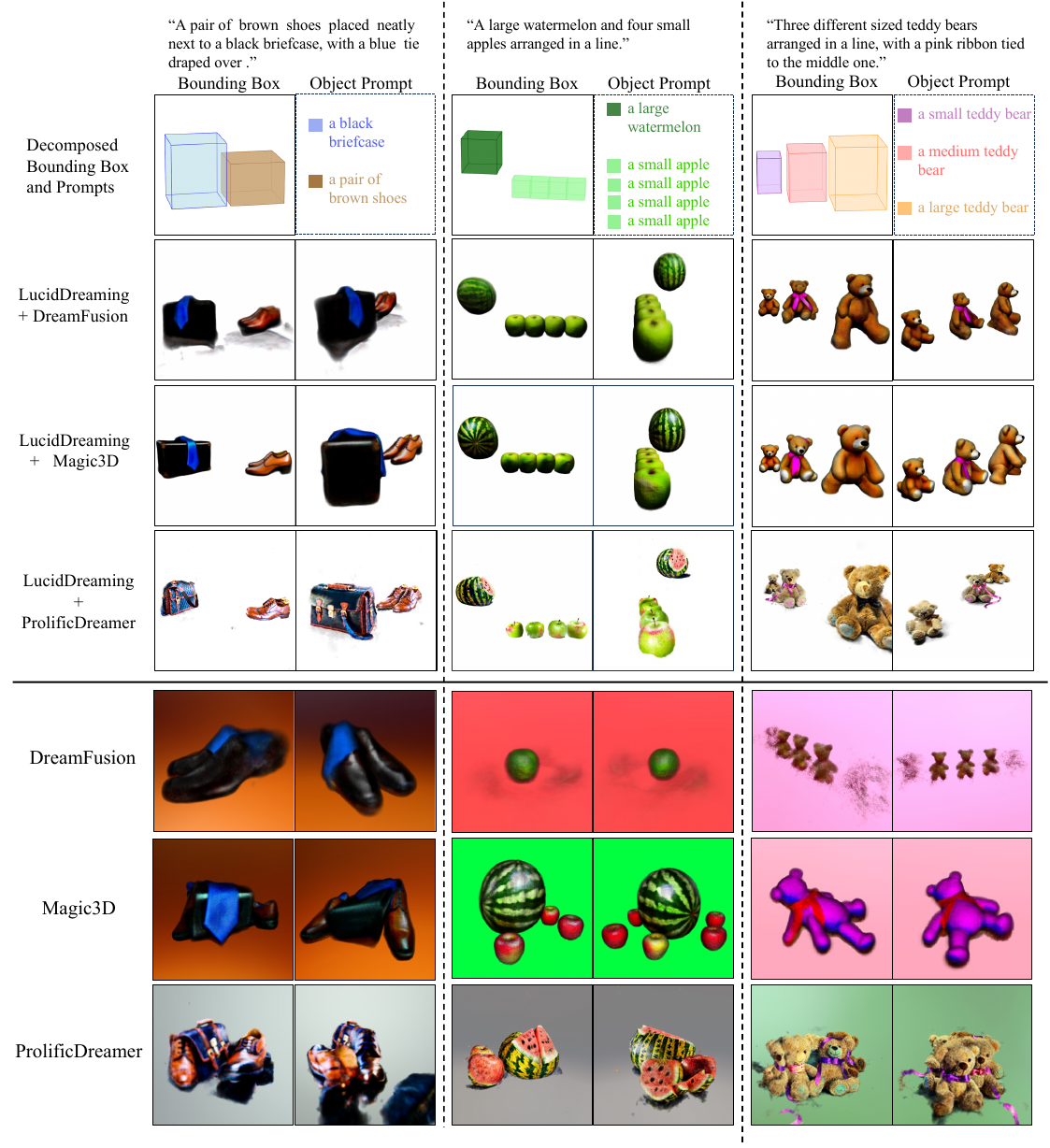}
    \caption{More visualization results of our \ourmodel compared to baseline methods. We show the 3D bounding box, text prompt, and two different views.}
    \label{fig:vis_3}
\end{figure*}

\begin{figure*}[!]
   \centering
    \includegraphics[width=\linewidth]{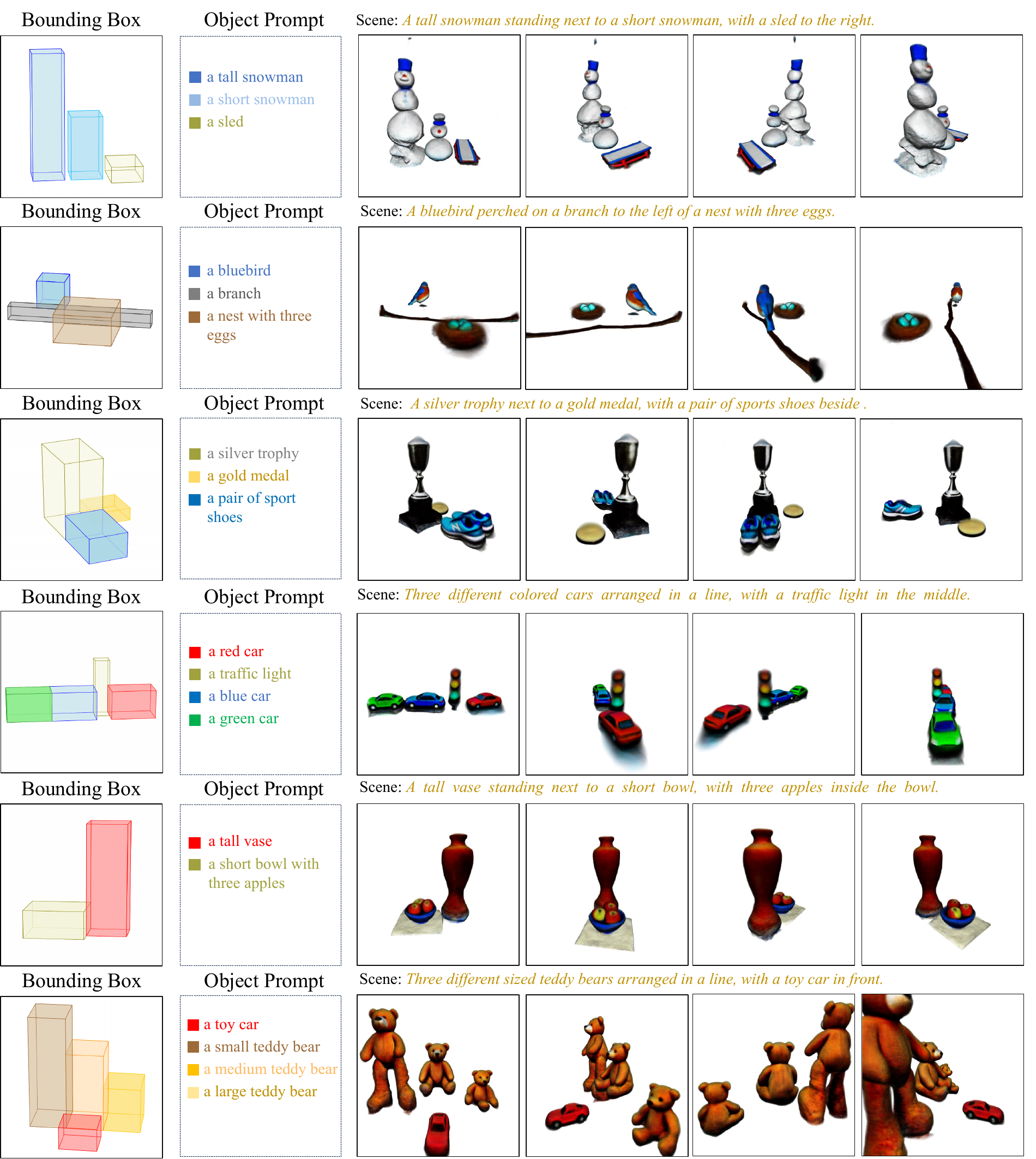}
    \caption{More visualization results of our \ourmodel with \textbf{DreamFusion}~\cite{poole2022dreamfusion} backbone.}
    \label{fig:more_vis_DF}
\end{figure*}

\begin{figure*}[!]
   \centering
    \includegraphics[width=\linewidth]{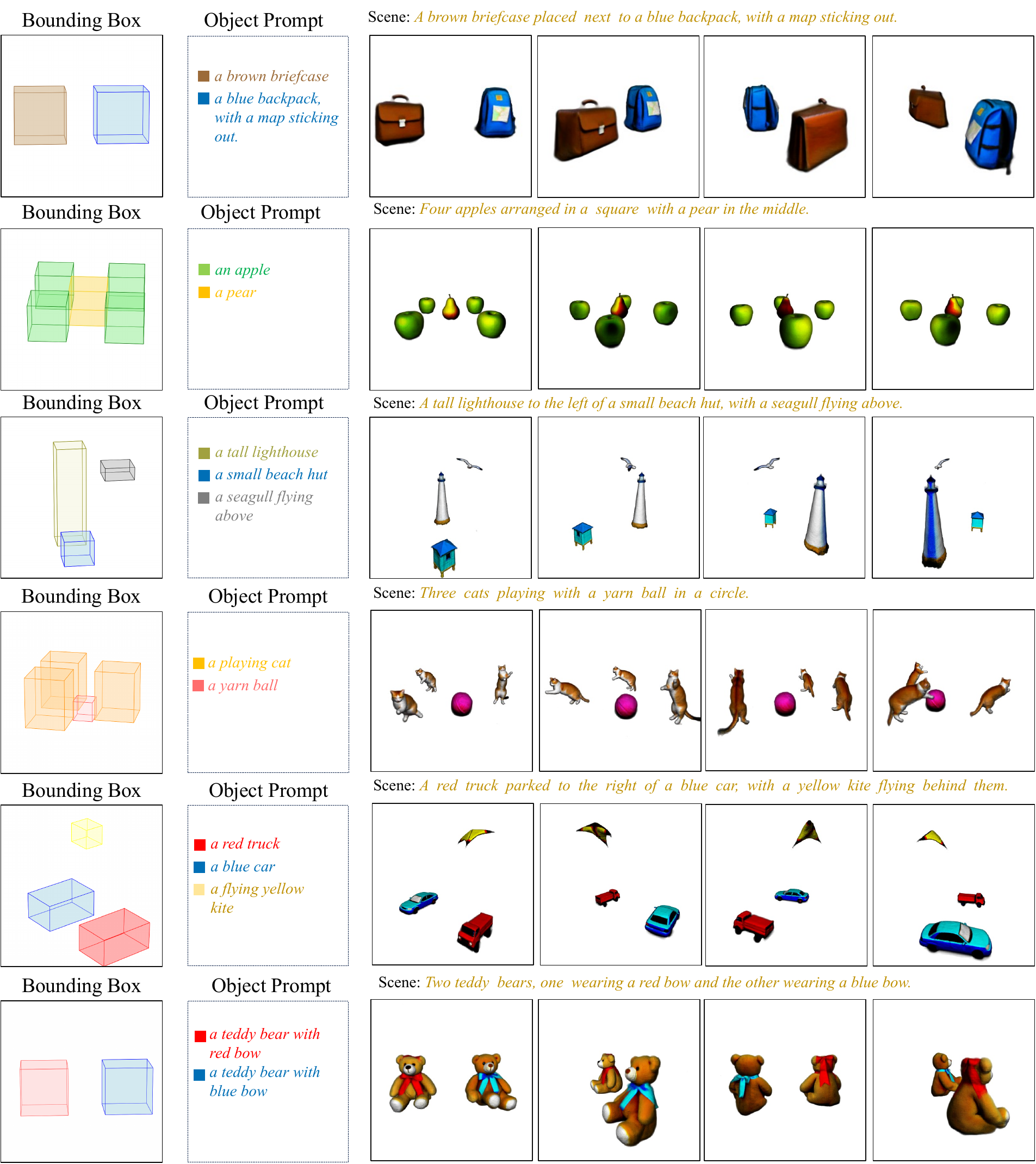}
    \caption{More visualization results of our \ourmodel with \textbf{Magic3D}~\cite{lin2023magic3d} backbone.}
    \label{fig:more_vis_m3d}
\end{figure*}

\begin{figure*}[!]
   \centering
    \includegraphics[width=\linewidth]{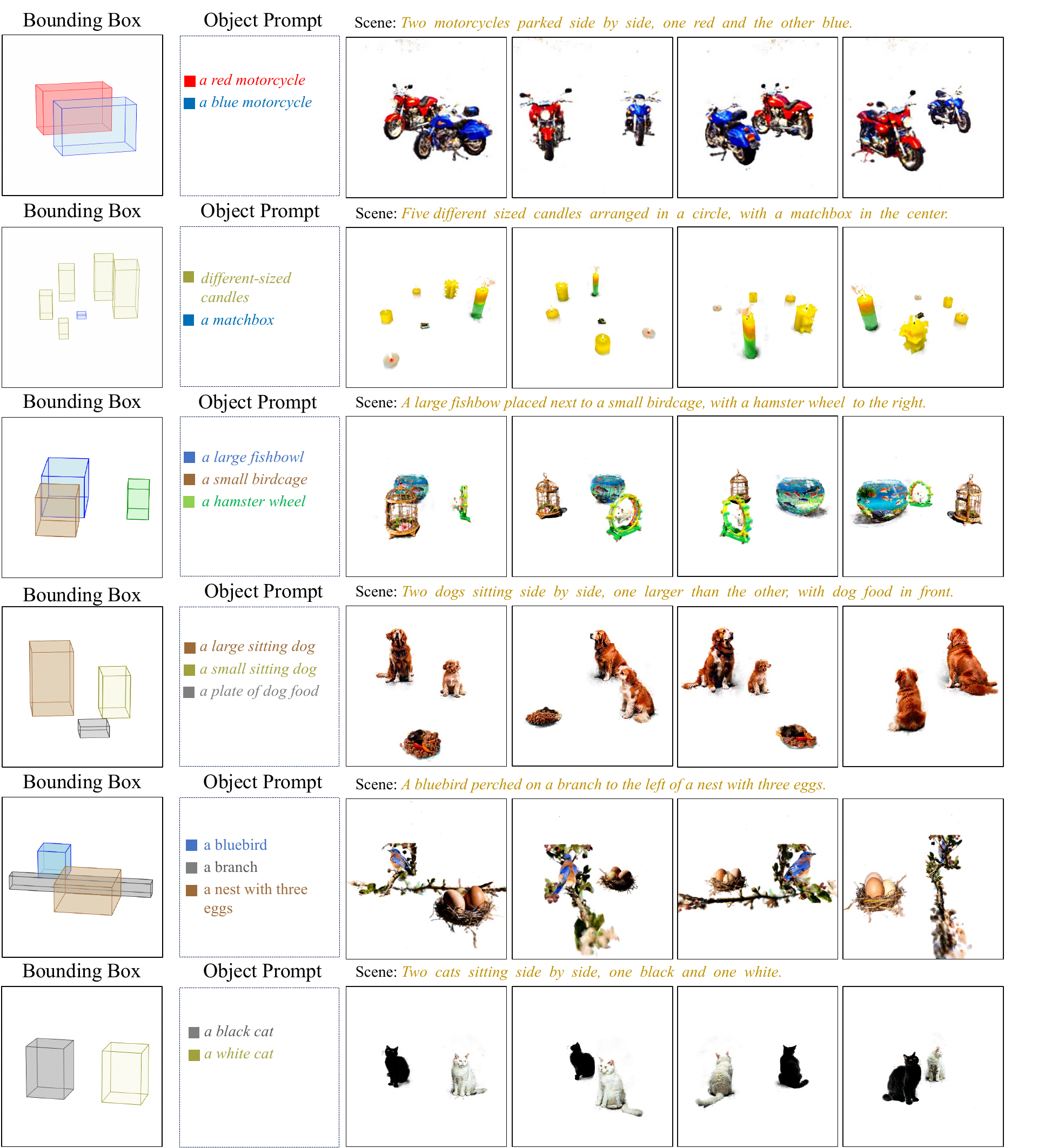}
    \caption{More visualization results of our \ourmodel with \textbf{ProlificDreamer}~\cite{wang2023prolificdreamer} backbone.}
    \label{fig:more_vis_pd}
\end{figure*}

Finally, we provide a full qualitative comparison with the baseline methods in Figs.~\ref{fig:vis_1},~\ref{fig:vis_2},~\ref{fig:vis_3}. It can clearly be seen that our method aligns the 3D generation with the bounding box while the baselines, using only text prompts as controlling conditions, fail in the wrong number of objects, object clustering/missing, and poor quality.  The result demonstrates the significant advantages of \ourmodel in terms of generation quality and controllability. We show more qualitative results in Figs.~\ref{fig:more_vis_DF},~\ref{fig:more_vis_m3d},~\ref{fig:more_vis_pd}.




\begin{table*}[h]
    \centering
    \begin{tabular}{c|p{0.9\textwidth}}
        \textbf{Type} & \textbf{Prompt} \\
        \hline
        System & You are my assistant to evaluate the correspondence of the image to a given text prompt.
Briefly describe the image within 50 words, focus on the objects in the image and their attributes (such as color, shape, texture), spatial layout and action relationships.

According to the image and your previous answer, evaluate how well the image aligns with the user's text prompt. Give a score from 0 to 100, according to the criteria:
\begin{itemize}[nosep]
    \item 100: The image perfectly matches the content of the text prompt, with no discrepancies.
    \item 80: The image portrayed most of the actions, events, and relationships but with minor discrepancies.
    \item 60: The image depicted some elements in the text prompt, but ignored some key parts or details.
    \item 40: The image did not depict any actions or events that match the text.
    \item 20: The image failed to convey the full scope in the text prompt.
\end{itemize}
Provide your analysis and explanation in JSON format with the following keys: score (e.g., 85), explanation (within 20 words).
 \\
        \hline
        Text & Please return the evaluation of the given image with text prompt: \textbf{xxx} \\
        \hline
        image & \textit{encoded image} \\
        
    \end{tabular}
    \caption{The GPT4-V prompt for evaluating images. The \textbf{xxx} is the scene description, and we encode the image into \textit{base-64} format.}
    \label{tab:gpt4-v-prompt}
\end{table*}

\section{Broader Impact and Limitation.}
\label{limitation}
In this paper, we propose \ourmodel, a revolutionary method for controllable, object-centric 3D generation. \ourmodel addresses the controllability limitations of existing works without suffering the generation quality, and can be seamlessly integrated into current 3D generation pipelines in a plug-and-play manner. Additionally, we introduce a dataset and corresponding benchmarks for controllable 3D generation, paving the way for future research. 

Though \ourmodel produces compelling results in controllability with various 3D generation frameworks, it still has several limitations. Rendering objects separately with clipped ray sampling, \ourmodel struggles to create interactions between objects. Implementing a global SDS should resolve such restrictions to some extent.
Additionally, like current controlling methods~\cite{po2023compositional}, our training time increases linearly with the number of objects, which could pose a potential issue when dealing with a large number of objects.  We consider addressing them as directions for future research. Overall, our work sets a standard in the dataset, evaluation metrics, and clear strategies for the development of controllable 3D generation, offering significant potential for the application and expansion of 3D generation with better conditioning alignment.

\clearpage

%
%
\clearpage
\bibliographystyle{splncs04}
\bibliography{main}
\end{document}